%% file: main.tex
\documentclass[10pt,twocolumn,letterpaper]{article}

\usepackage[pagenumbers]{cvpr} 
\usepackage{multirow}
\usepackage{amsthm,amsmath}
\usepackage{mathtools}
\newtheorem{theorem}{Theorem}

\usepackage{relsize}
\usepackage{array,booktabs,multirow}
\usepackage{xcolor}
\usepackage{colortbl} 
\usepackage{wrapfig} 
\usepackage[accsupp]{axessibility}
\definecolor{eceLow}{HTML}{76B7B2}
\definecolor{eceMid}{HTML}{BAB0AC}
\definecolor{eceHigh}{HTML}{E15759}
\definecolor{eceLine}{HTML}{5B2A86}
\definecolor{regColor}{HTML}{A9D18E}
\definecolor{zsColor}{HTML}{26547C}
\definecolor{oursColor}{HTML}{E86AA3}


\definecolor{cvprblue}{rgb}{0.21,0.49,0.74}
\usepackage[pagebackref,breaklinks,colorlinks,allcolors=cvprblue]{hyperref}

\def\paperID{} 
\def\confName{CVPR}
\def\confYear{2026}
\title{Improving Calibration in Test-Time Prompt Tuning for Vision-Language Models via Data-Free Flatness-Aware Prompt Pretraining}

\author{
 Hyeonseo Jang\textsuperscript{1} \qquad
 Jaebyeong Jeon\textsuperscript{1} \qquad
 Joong-Won Hwang\textsuperscript{2} \qquad
 Kibok Lee\textsuperscript{1} \\
 \textsuperscript{1}Yonsei University \qquad
 \textsuperscript{2}ETRI \\
 {\tt\small \{jhyeonseo715, jaebyeong98, kibok\}@yonsei.ac.kr \quad jwhwang@etri.re.kr}
}

\begin{document}
\maketitle
\input{sec/0_abstract}    
\input{sec/1_Introduction}

\input{sec/2_RelatedWorks}

\input{sec/3_Preliminaries}

\input{sec/4_Analysis}
\input{sec/5_Methods}
\input{sec/6_Experiments}
\input{sec/7_Conclusion}

\input{sec/8_Acknowledgment}
{
    \small
    \bibliographystyle{ieeenat_fullname}
    \bibliography{main}
}

\input{sec/X_suppl}

\end{document}

%% file: sec/0_abstract.tex
\begin{abstract}
Test-time prompt tuning (TPT) has emerged as a promising technique for enhancing the adaptability of vision-language models by optimizing textual prompts using unlabeled test data.
However, prior studies have observed that TPT often produces poorly calibrated models, raising concerns about the reliability of their predictions.
Recent works address this issue by incorporating additional regularization terms that constrain model outputs, which improve calibration but often degrade performance.
In this work, we reveal that these regularization strategies implicitly encourage optimization toward flatter minima, and that the sharpness of the loss landscape around adapted prompts is a key factor governing calibration quality.
Motivated by this observation, we introduce Flatness-aware Prompt Pretraining (FPP), a simple yet effective pretraining framework for TPT that initializes prompts within flatter regions of the loss landscape prior to adaptation.
We show that simply replacing the initialization in existing TPT pipelines---without modifying any other components---is sufficient to improve both calibration and performance.
Notably, FPP requires no labeled data and incurs no additional computational costs during test-time tuning, making it highly practical for real-world deployment.
The code is available at: \url{https://github.com/YonseiML/fpp}.
\end{abstract}

%% file: sec/1_Introduction.tex
\section{Introduction}
\label{sec:Introduction}

Vision-language models such as CLIP~\cite{Radford2021} have recently achieved remarkable zero-shot performance across a wide range of downstream tasks~\cite{Jia2021, Radford2021}. 
In these models, textual input templates play a crucial role in determining performance, motivating recent advances in prompt tuning methods that optimize prompt templates in a few-shot manner~\cite{Zhou2022b, Zhou2022a, Khattak2023a, Khattak2023b}.
However, these approaches often face limitations in real-world applications, as they cannot directly adapt to distribution shifts in target tasks~\cite{Wang2024, Shu2022}.
To address this limitation, test-time prompt tuning (TPT)~\cite{Shu2022} introduces an entropy minimization (EM) loss that enables prompt adaptation without requiring labeled data, inspiring extensive subsequent research~\cite{Feng2023, Zhu2024, Xiao2025, Sharifdeen2025, Yoon2024}.

\begin{figure}[t!]
\begin{center}
    \includegraphics[width=0.86\columnwidth]{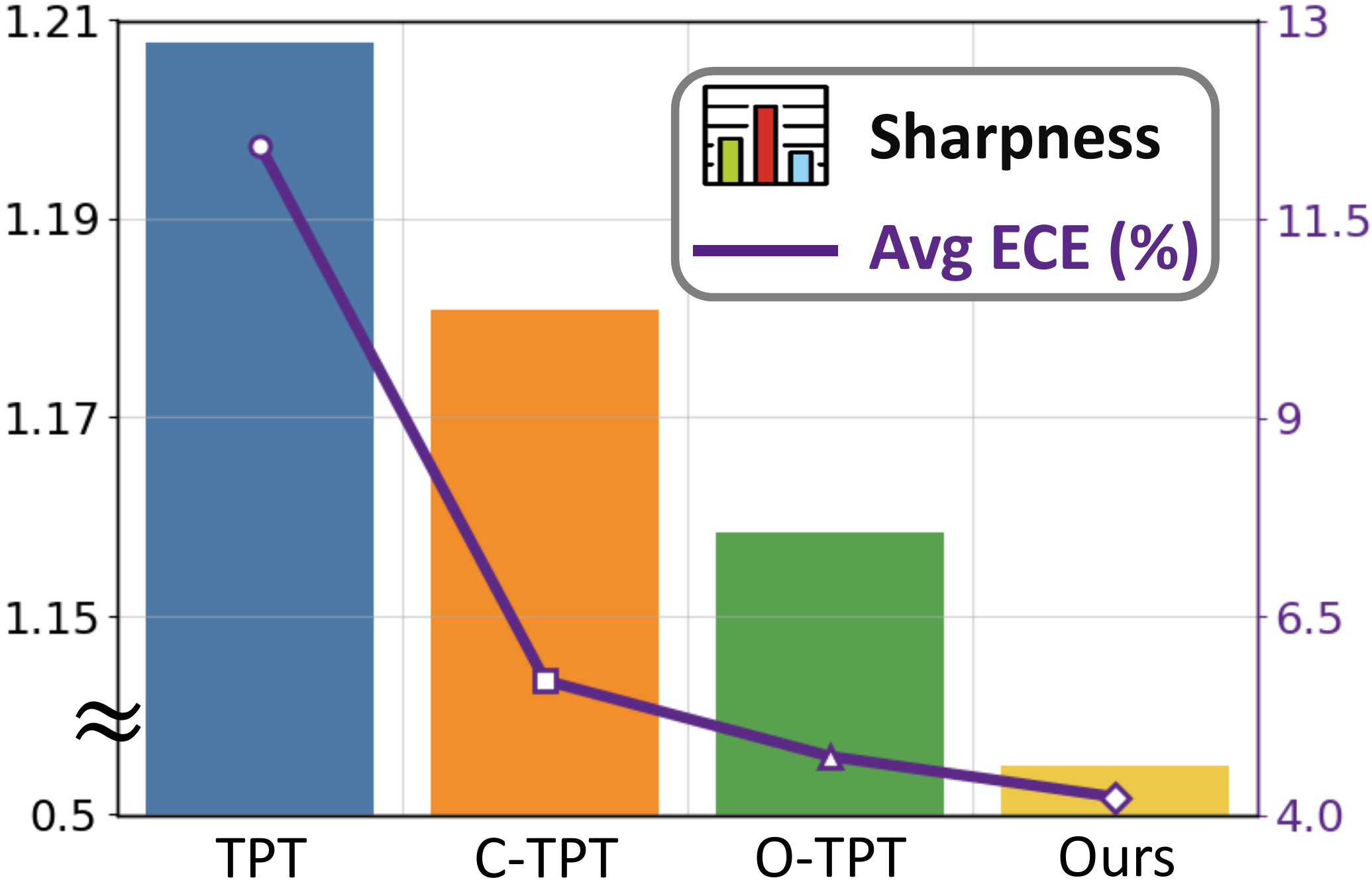}
    \vspace{-8pt}
    \caption{Applying regularization loss into TPT (C-TPT and O-TPT) encourages the prompt to converge toward a flatter loss landscape, leading to reduced Expected Calibration Error (ECE).}
    \label{fig:flatness}
\vspace{-26pt}
\end{center}
\end{figure}

However, TPT-based methods have been observed to degrade model calibration, leading to a mismatch between prediction confidence and actual accuracy~\cite{Murugesan2024, Yoon2024, Sharifdeen2025, Basu2025}.
Since VLMs are widely deployed in domains that require reliable uncertainty estimates, such as healthcare~\cite{Wang2022, Liu2023} and autonomous systems~\cite{Khandelwal2022, Bucker2022}, ensuring proper calibration in TPT settings has emerged as a critical research direction.
Recent studies primarily address this issue by regularizing TPT with additional loss terms~\cite{Yoon2024, Sharifdeen2025, Basu2025}, which improve calibration through geometric constraints that encourage broader dispersion of output text features.
Although these approaches demonstrate empirical success, they offer limited insight into how such regularized optimization affects the learned input prompts, leaving the mechanisms underlying calibration improvement largely unexplored.

In this paper, we show that existing regularization strategies for TPT implicitly act as mechanisms that guide prompts toward flat minima in the loss landscape, which are known to improve generalization in neural networks~\cite{Dinh2017, Li2022, Zhou2022, Keskar2017}.
As illustrated in Fig.~\ref{fig:flatness}, prompts optimized via regularized TPT exhibit reduced sharpness in their surrounding loss landscapes, accompanied by a corresponding decrease in expected calibration error (ECE).
Notably, we further observe that adapted prompts lying in flat minima consistently achieve substantially lower calibration errors than those trapped in sharp minima, indicating that convergence to flatter regions of the loss landscape is associated with improved calibration.
These findings highlight that prompt tuning within flat minima is crucial to achieve effective calibration. 
However, explicitly enforcing flatness through geometric regularization of output features often degrades performance.
Furthermore,  existing methods~\cite{Yoon2024, Sharifdeen2025, Basu2025} struggle to adequately explore flatter regions of the loss landscape, as increasing the number of iterations also incurs the risk of overconfidence in incorrect predictions.

Building on these insights, we propose FPP, a \textbf{F}latness-aware \textbf{P}rompt \textbf{P}retraining framework that initializes prompts within flatter regions of the loss landscape before TPT. 
Our FPP employs two complementary objectives: (i) an alignment loss that encourages the learned prompts to generate text features consistent with those of the original prompts, and (ii) a flatness loss that reduces output sensitivity to random perturbations.
By jointly optimizing these objectives, FPP generates new initial prompts with provably reduced sharpness for any differentiable loss function, facilitating efficient convergence toward flat minima even in a single step.
Because FPP serves as a pretraining scheme for prompt initialization, it can be seamlessly integrated with existing TPT-based methods by replacing their initial prompts with our pretrained ones, without altering subsequent adaptation procedures.
Moreover, FPP operates without relying on any external resources such as training or testing image samples, and incurs no extra computational cost during adaptation, making it practical for real-world deployment.
Our contributions are summarized as follows:

\begin{itemize}
    \item We identify a strong correlation between flat minima and calibration quality, and further reveal that existing calibration methods for TPT implicitly guide prompts toward flatter regions of the loss landscape.
    \item We introduce FPP, a flatness-aware prompt pretraining framework for TPT that initializes prompts in flatter regions of the loss landscape without relying on additional resources, enabling effective convergence to flat minima.
    \item Our extensive experiments demonstrate that simply initializing existing TPT-based methods with our pretrained prompts achieves state-of-the-art (SOTA) results in both calibration and performance in downstream tasks.
\end{itemize}

%% file: sec/2_RelatedWorks.tex
\section{Related Work}
\label{sec:Related Works}

\noindent\textbf{Prompt Tuning.} 
The zero-shot performance of CLIP~\cite{Radford2021} is highly sensitive to the phrasing of manually crafted input textual prompts, where even slight variations can lead to substantial changes in accuracy~\cite{Zhou2022b}.
To address this issue, prompt tuning methods such as CoOp~\cite{Zhou2022b} and CoCoOp~\cite{Zhou2022a} replace fixed textual templates with learnable prompt vectors that can be trained through few-shot learning.
Building on this idea, MaPLe~\cite{Khattak2023a} introduces a multi-modal approach that generates visual prompts from text prompts via linear projections.
Another line of work, TPT~\cite{Shu2022}, extends the test-time adaptation (TTA) paradigm to prompt tuning by employing an EM objective, enabling unsupervised adaptation to distribution shifts in test data.
DiffTPT~\cite{Feng2023} further enhances this approach by incorporating diffusion-generated images to improve accuracy.
More recently, Self-TPT~\cite{Zhu2024} proposed a method that further adapts the few-shot learned prompts to class names before test-time, thereby improving generalization across classes.
Despite their impressive improvement in accuracy, these methods are designed without considering model calibration, often resulting in overconfident predictions~\cite{Yoon2024, Sharifdeen2025}.

\noindent\textbf{Calibration of Neural Networks.} 
Calibration measures how well the predicted confidence of a model aligns with its actual accuracy~\cite{Guo2017}. 
Broadly, calibration approaches fall into two categories: post-hoc~\cite{Guo2017, Platt1999, Vovk2005, Lei2018} and train-time~\cite{Kumar2018, Karandikar2021, Yoon2023} methods.
Post-hoc methods, such as temperature scaling~\cite{Guo2017} and Platt scaling~\cite{Platt1999}, adjust a trained model using a held-out validation set to better match predicted probabilities with observed outcomes. While effective, these techniques depend on labeled datasets that closely resemble the target distribution, limiting their flexible application~\cite{Liu2022b}.
Train-time calibration, on the other hand, incorporates additional calibration objectives during model training to encourage predictions that better align with true probabilities.
In the TTA setting, C-TPT~\cite{Yoon2024} and O-TPT~\cite{Sharifdeen2025} follow this paradigm by incorporating an additional regularization loss term to improve calibration.

\noindent\textbf{Flat Minima.} 
The relationship between sharpness of the loss landscape and generalization has been widely explored, showing that flatter local minima improve model generalization~\cite{Dinh2017, Li2022, Zhou2022, Keskar2017}.
In this regard, SAM~\cite{Foret2021} and its subsequent studies~\cite{Zhuang2022, Liu2022a, Kwon2021, Li2024} introduce a new optimization objective that seeks flat minima by uniformly minimizing the loss within a neighborhood. 
Following them, some TTA methods~\cite{Niu2023, Gong2023} incorporate the SAM objective into the existing EM loss, leveraging flatter loss landscapes to suppress the influence of noisy samples while continuously updating model parameters. 
However, despite extensive research on flat minima, their connection to calibration remains largely unexplored and somewhat controversial.
For example,~\cite{Williams2024} report that some regularization techniques, such as data augmentation, improve calibration but often yield sharper minima.
Conversely, CSAM~\cite{Tan2025} shows that the SAM objective acts as an implicit entropy regularizer and leads to improved calibration.
However, these studies primarily focus on supervised settings and emphasize the effects of regularization on calibration rather than examining the intrinsic properties of flat minima themselves.
In contrast, we demonstrate that even in the absence of explicit regularization, flat minima inherently contribute to better calibration.

%% file: sec/3_Preliminaries.tex
\section{Preliminaries}
\label{sec:Preliminaries}

\subsection{Prompt Tuning for CLIP at Test-Time}
\noindent\textbf{CLIP-Based Classification.} 
Given an image $X$, a set of $K$ class names $C=\{c_1, c_2, \ldots, c_K\}$, and a textual prompt~$\theta$, CLIP encodes them into a joint embedding space using an image encoder $f_I(\cdot)$ and a text encoder $f_T(\cdot)$.
The image feature is represented as $v=f_I(X)$, while the text feature for class $c_k$ is $t_k=f_T(c_k, \theta)$.
Class scores are computed as cosine similarities $\cos(v, t_k)$, and subsequently converted into probabilities using a temperature-scaled softmax function with temperature $\tau$.
The predicted class is given by $\hat{c} = \arg\max_{c_k} P(c_k|\theta,v)$, with confidence $\hat{P} = \max_{c_k} P(c_k|\theta,v)$.

\vspace{2pt}\noindent\textbf{Test-Time Prompt Tuning (TPT).} 
TPT~\cite{Shu2022} performs sample-specific adaptation by optimizing a separate prompt $\theta$ for each test pair $(X, C)$ using an entropy minimization (EM) objective. 
In this setting, the class name set $C$ is fixed, while the image sample $X$ varies at each step, being provided sequentially.
The EM loss is computed from a set of image features $V$, obtained from high-confidence augmented views of $X$, and is defined as:
\begin{equation}
{L}_{\text{ent}}(\theta, V) 
=  -\sum_{k=1}^K P(c_k\mid \theta, V) 
\log P(c_k\mid \theta, V),
\label{eq:entropy_loss}
\end{equation}
where $P(c_k \mid \theta, V)$ denotes the prediction probability for class $c_k$, averaged across the augmented views.

\vspace{2pt}\noindent\textbf{Regularized TPT.} 
Motivated by the empirical observation that higher dispersion of text features is associated with lower calibration error, C-TPT~\cite{Yoon2024} and O-TPT~\cite{Sharifdeen2025} incorporate an additional regularization term, $L_\text{reg}$, to explicitly encourage feature diversity:
\vspace{-8pt}
\begin{align}
L_{\text{reg}}^{\text{C-TPT}}(\theta) &= -\dfrac{1}{K} \sum_{k=1}^{K} \| t_k - \mu \|_2, 
\label{eq:ctpt_detail} \\
L_{\text{reg}}^{\text{O-TPT}}(\theta) &= \| T T^{\top} - I_K \|_2^2,
\label{eq:otpt_detail}
\end{align}
where $T = f_T(C; \theta) = [t_1, \dots, t_K]^\top$ denotes the text feature matrix collecting the $K$ text features, $\mu = \tfrac{1}{K}\sum_{k=1}^K t_k$ is the mean text feature vector, and $I_K$ is the identity matrix.
For brevity, we omit the dependence on $C$ in the above equations, as it remains fixed for each dataset. 
The overall objective combines the EM loss with the regularization term, weighted by a hyperparameter $\lambda$:
\begin{equation}
{L}_{\text{total}}(\theta, V) 
= {L}_{\text{ent}}(\theta, V) 
+ \lambda {L}_{\text{reg}}(\theta).
\label{eq:total_loss}
\end{equation}

\subsection{Sharpness-Aware Minimization (SAM)}
In prior works~\cite{Foret2021, Zhuang2022, Kwon2021}, \textit{sharpness of the loss landscape} is commonly quantified by the maximum loss difference between the current parameter $w_t$ and its perturbed counterpart.
Formally, this measure is defined as follows: 
\begin{align}
h(w_t, v) &\triangleq \max_{\|\varepsilon\|\le \rho} L(w_t+\varepsilon,v) - L(w_t,v),
\label{eq:sharpness}
\end{align}
where $\rho$ controls the magnitude of the perturbation.

To identify flat minima in the loss landscape and achieve better generalization, recent studies~\cite{Foret2021, Kwon2021, Liu2022a, Zhuang2022} minimize this sharpness using a common two-step optimization procedure.
Starting from the current parameter $w_t$, they first search for a nearby perturbation $\varepsilon$ such that $w_t+\varepsilon$ yields a higher loss. 
The gradient is then computed at this perturbed point and used to update $w_t$.
This procedure encourages the model to reduce loss not only at $w_t$ but throughout its neighborhood, effectively guiding parameters toward flatter regions of the loss landscape~\cite{Foret2021}.

A representative example is Sharpness-Aware Minimization (SAM)~\cite{Foret2021}, which seeks the perturbation that maximizes the loss within a local neighborhood.
Because exactly determining such a perturbation is computationally intractable~\cite{Liu2022a}, SAM approximates it using a first-order Taylor expansion, leading to the following update equation:
\begin{equation}
\label{eq:sam}
w_{t+1}
= w_{t} - \eta\,\left.\nabla_w L(w,v)\right|_{ w_{t}+\hat\varepsilon_{t}}.
\end{equation}
Here, $\hat\varepsilon_{t}=\rho\cdot\frac{\nabla_w L(w_t,v)}{|\nabla_w L(w_t,v)|}$ denotes the approximated perturbation.
Following prior works~\cite{Foret2021, Zhuang2022, Kwon2021}, we adopt this perturbation form when computing the sharpness defined in Eq.~\eqref{eq:sharpness} throughout our experiments.

%% file: sec/4_Analysis.tex
\section{Analysis of Regularized TPT through SAM}
\label{sec:Analysis}

In this section, we show that promoting higher dispersion of text features via regularization guides prompts toward flat minima in TPT frameworks.
We begin by revisiting the standard setup of TPT-based methods~\cite{Shu2022, Yoon2024, Sharifdeen2025, Feng2023}.
In these approaches, the learned prompt is reinitialized to a predefined prompt $\theta_{0}^\text{zs}$ for every test sample.
Thus, for a fixed set of class names $C$, adaptation always starts from the same text features, denoted as $f_T(C, \theta_{0}^\text{zs})$.
Consequently, since the regularization terms in Eq.~\eqref{eq:ctpt_detail} and Eq.~\eqref{eq:otpt_detail} depend solely on these features, their gradients $\nabla_{\theta} L_\text{reg}(\theta)\big|_{\theta_{0}^\text{zs}}$ remain identical across all test samples.

Because the methods we analyze perform a single-step update per sample~\cite{Yoon2024, Sharifdeen2025}, accumulated effects such as momentum can be ignored.
For analytical clarity, we adopt a simple gradient-based update rule and assume a learning rate set to one.
Consequently, the update formulation derived from Eq.~\eqref{eq:total_loss} is given by:
\begin{align}
\theta_1^{\text{reg}}
&= \theta_{0}^{\text{zs}}
- \left.\nabla_{\theta} L_{\text{ent}}(\theta,V)\right|_{\theta_{0}^{\text{zs}}}
- \lambda\,\left.\nabla_{\theta} L_{\text{reg}}(\theta)\right|_{\theta_{0}^{\text{zs}}} \nonumber\\
&= \theta_{0}^{\text{zs}}
- \left.\nabla_{\theta} L_{\text{ent}}(\theta,V)\right|_{\theta_{0}^{\text{zs}}}
- \varepsilon_\text{reg},
\label{eq:update_equation_full}
\end{align}
where $\varepsilon_{\text{reg}}$ denotes the constant offset induced by the gradient of the regularization loss, and subscripts $0$ and $1$ denote prompts before and after the update, respectively.

By combining the two fixed terms in Eq.~\eqref{eq:update_equation_full}, we define $\theta_{0}^\text{reg} := \theta_{0}^\text{zs} - \varepsilon_\text{reg}$ as a new initialization for TPT. 
Under this definition, the regularized TPT becomes equivalent to the original TPT  formulation without the regularization term: 
\begin{align}
\theta_{1}^\text{reg}
&= \theta_{0}^\text{reg}
- \left.\nabla_{\theta} L_\text{ent}(\theta,V)\right|_{\theta_{0}^\text{zs}} \nonumber\\
&= \theta_{0}^\text{reg}
- \left.\nabla_{\theta} L_\text{ent}(\theta,V)\right|_{\theta_{0}^\text{reg}+\varepsilon_\text{reg}}.
\label{eq:update_equation}
\end{align}
Here, the gradient of the EM loss is always computed at the perturbed point $\theta_{0}^\text{reg} + \varepsilon_\text{reg}$ rather than at the point where it is ultimately applied, $\theta_{0}^\text{reg}$.

\begin{figure}[t!]
\begin{center}
    \includegraphics[width=0.98\columnwidth]{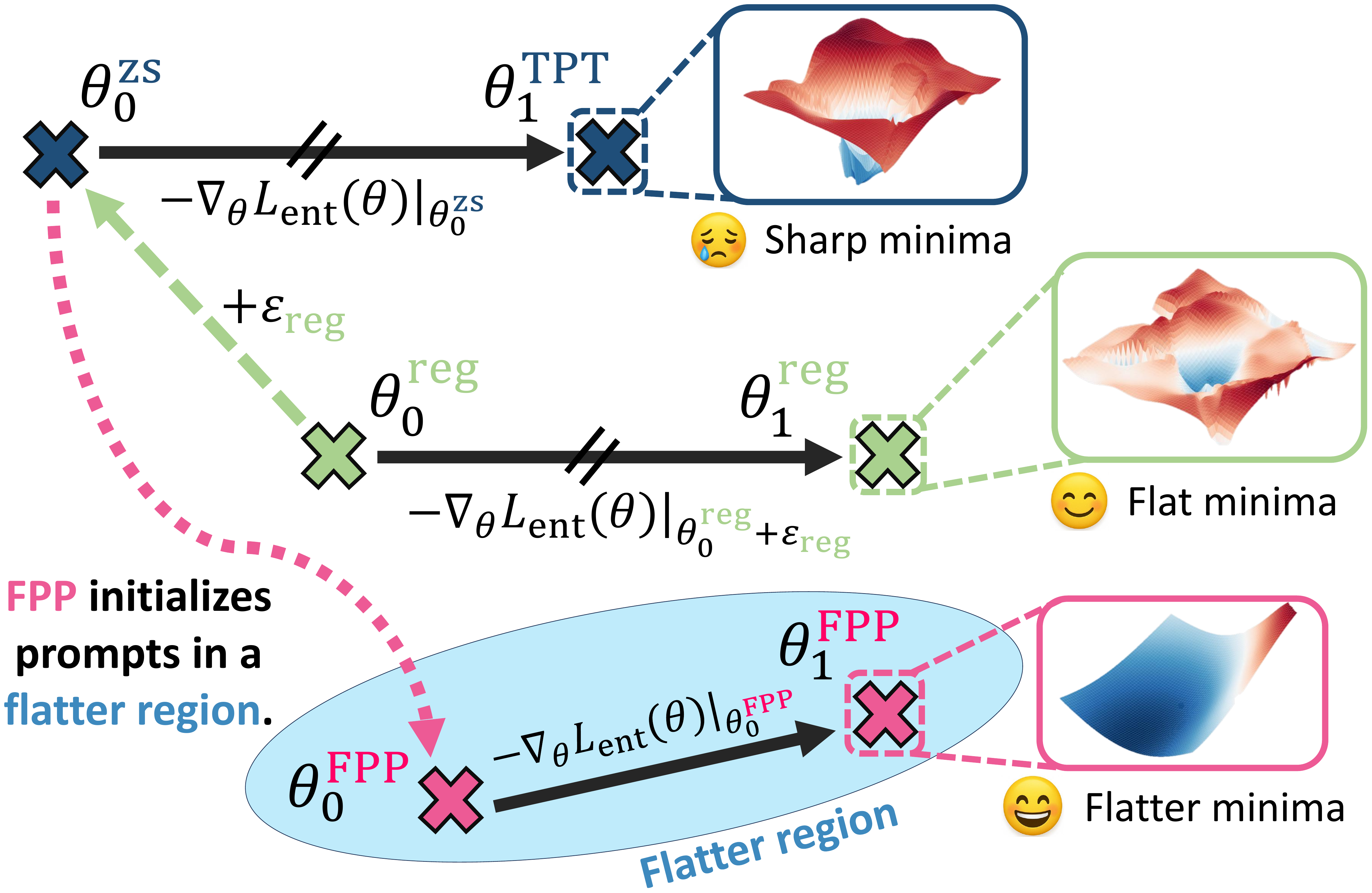}
    \vspace{-5pt}
    \caption{Regularized TPT can be interpreted as an optimization in which the initial prompt is shifted and fixed at $\theta_0^{\text{\textcolor{regColor}{reg}}}$, while gradients of the EM losses are computed at a perturbed point $\theta_0^{\text{\textcolor{zsColor}{zs}}}$. This mechanism encourages convergence toward flat minima. In contrast, our \textcolor{oursColor}{FPP} pretrains the initial prompt to reside in a flatter region, facilitating convergence to flat minima without perturbation.}
    \label{fig:gradient_update}
\vspace{-16pt}
\end{center}
\end{figure}

Notably, because $\varepsilon_\text{reg}$ corresponds to the gradient of the regularization loss, the perturbed point $\theta_{0}^\text{reg} + \varepsilon_\text{reg}$ yields a higher regularization loss than the original point $\theta_{0}^\text{reg}$.
This encourages text features from different classes to cluster more closely.
In other words, adding $\varepsilon_\text{reg}$ can be interpreted as introducing a perturbation that produces more uniform prediction probabilities, which corresponds to an increase in the EM loss.
The following theorem formalizes the connection between the regularization objective and the EM loss (see Appendix~\ref{sec:Theory_1} for the full statement and proof):
\begin{theorem}[Informal]\label{thm:reg_ent_corr}
Let \(\mathbb S^{D-1} = \{ v \in \mathbb{R}^{D}: \lVert v \rVert_2 = 1 \} \) be the unit \((D-1)\)-sphere, where \(D\) is the dimension of the feature space.
For an image feature \(v\) sampled from the uniform distribution on \(\mathbb S^{D-1}\), the expected EM loss is
\[
\mathcal H(T):=\mathbb E_{v\sim \mathrm{Unif}(\mathbb S^{D-1})}\!\left[L_{\mathrm{ent}}(T,v)\right].
\]
Then, for constants \(\alpha>0\) and \(\beta\),
\(L_{\mathrm{reg}}\) corresponding to Eq.~\eqref{eq:ctpt_detail} and Eq.~\eqref{eq:otpt_detail} satisfy
\[
\mathcal H(T)=\alpha L_{\mathrm{reg}}(T)+\beta+O(D^{-3/2}).
\]
Therefore, for sufficiently large \(D\), increasing the regularization loss also leads to increase the expected EM loss.
\end{theorem}
Recall that, in regularized TPT, $\theta_{0}^\text{reg}$ is updated in the direction of a gradient computed at a perturbed point $\theta_{0}^\text{reg} + \varepsilon_\text{reg}$, as shown in Eq.~\eqref{eq:update_equation}.
According to Theorem 1, this perturbed point is likely to yield a higher EM loss than the original point.
This mechanism aligns with the core principle of the SAM in Eq.~\eqref{eq:sam}, where gradients are computed at perturbed points that yield higher loss within the neighborhood.
Fig.~\ref{fig:flatness} empirically confirms that the regularization term effectively reduces sharpness, guiding the prompts toward flat minima.
Fig.~\ref{fig:gradient_update} illustrates this mechanism.

%% file: sec/5_Methods.tex
\section{Flatness-Aware Prompt Pretraining (FPP)}
\label{sec:Methods}

\begin{table*}[t!]
\centering
\resizebox{0.95\textwidth}{!}{%
\begin{tabular}{l|c|c|cccccccccc|c}
\toprule
\textbf{Method} & \textbf{Venue} & \textbf{Metric} & \textbf{Air} & \textbf{Calt} & \textbf{Car} & \textbf{DTD} & \textbf{SAT} & \textbf{FLW} & \textbf{Food} & \textbf{Pets} & \textbf{SUN} & \textbf{UCF} & \textbf{Avg.} \\
\midrule
\multirow{3}{*}{CLIP~\cite{Radford2021}} & \multirow{3}{*}{ICML 2021} & Acc. &  23.9 & 92.9 & 65.3 & 44.3 & 41.3 & 67.3 & 83.6 & 88.0 & 62.5 & 65.0 & 63.41 \\
                           &                      & ECE & 5.11 & 5.50 & 4.25 & 8.50 & 7.40 & 3.00 & 2.39 & 4.37 & 2.53 & 3.59 & 4.67 \\
                           &                      & SCE & 0.52 & 0.25 & 0.23 & 1.33 & 6.18 & 0.59 & 0.20 & 0.68 & 0.12 & 0.52 & 1.06 \\
\midrule
\multirow{3}{*}{TPT~\cite{Shu2022}}       & \multirow{3}{*}{NeurIPS 2022} & Acc. &  23.4 & 93.8 & 66.3 & 46.7 & 42.4 & 69.0 & 84.7 & 87.1 & 65.5 & 67.3 & \underline{64.62} \\
                           &                      & ECE & 16.8 & 4.51 & 5.16 & 21.2 & 21.5 & 13.5 & 3.98 & 5.77 & 11.3 & 13.0 & 11.67 \\
                           &                      & SCE & 0.58 & 0.16 & 0.25 & 1.44 & 7.07 & 0.51 & 0.17 & 0.60 & 0.15 & 0.57 & 1.15 \\
\midrule
\multirow{3}{*}{C-TPT~\cite{Yoon2024}}     & \multirow{3}{*}{ICLR 2024} & Acc. &  24.0 & 93.6 & 65.8 & 46.0 & 43.2 & 69.8 & 83.7 & 88.2 & 64.8 & 65.7 & 64.48 \\
                           &                            & ECE & 4.36 & 4.24 & 1.59 & 11.9 & 13.2 & 5.04 & 3.43 & 1.90 & 5.04 & 2.54 & 5.32 \\
                           &                            & SCE & 0.56 & 0.22 & 0.22 & 1.31 & 6.81 & 0.52 & 0.22 & 0.58 & 0.14 & 0.52 & \underline{1.11} \\
\midrule
\multirow{3}{*}{O-TPT~\cite{Sharifdeen2025}}     & \multirow{3}{*}{CVPR 2025} & Acc. & 23.64 & 93.95 & 64.53 & 45.68 & 42.84 & 70.07 & 84.13 & 87.95 & 64.23 & 64.16 & 64.12 \\
                           &                           & ECE & 3.68 & 3.80 & 1.78 & 7.88 & 12.98 & 3.87 & 1.46 & 1.90 & 4.93 & 2.34 & \underline{4.46} \\
                           &                           & SCE & 0.56 & 0.17 & 1.07 & 1.24 & 6.58 & 0.53 & 0.19 & 0.57 & 0.12 & 0.51 & 1.15 \\
\midrule
\cellcolor{gray!20}           & \cellcolor{gray!20}    & \cellcolor{gray!20}Acc. &  \cellcolor{gray!20}24.75 &
                              \cellcolor{gray!20}93.25 & \cellcolor{gray!20}66.65 & \cellcolor{gray!20}46.14 & \cellcolor{gray!20}50.66 & \cellcolor{gray!20}69.43 & \cellcolor{gray!20}84.31 & \cellcolor{gray!20}87.30 & \cellcolor{gray!20}64.08 & \cellcolor{gray!20}67.08 & \cellcolor{gray!20}\textbf{65.37} \\
                              \cellcolor{gray!20}
                              & \cellcolor{gray!20}   & \cellcolor{gray!20}ECE  &  \cellcolor{gray!20}7.26 & \cellcolor{gray!20}5.85 & \cellcolor{gray!20}2.00 & \cellcolor{gray!20}7.52 & \cellcolor{gray!20}5.19 & \cellcolor{gray!20}2.67 & \cellcolor{gray!20}1.92 & \cellcolor{gray!20}2.63 & \cellcolor{gray!20}3.22 & \cellcolor{gray!20}3.04 & \cellcolor{gray!20}\textbf{4.13} \\
\multirow{-3}{*}{\cellcolor{gray!20}FPP (Ours)}   & \multicolumn{1}{c|}{\multirow{-3}{*}{\cellcolor{gray!20}CVPR 2026}}    & \cellcolor{gray!20}SCE  &  \cellcolor{gray!20}0.57 & \cellcolor{gray!20}0.26 & \cellcolor{gray!20}0.22 & \cellcolor{gray!20}1.31 & \cellcolor{gray!20}5.12 & \cellcolor{gray!20}0.58 & \cellcolor{gray!20}0.20 & \cellcolor{gray!20}0.70 & \cellcolor{gray!20}0.12 & \cellcolor{gray!20}0.51 & \cellcolor{gray!20}\textbf{0.96} \\
\bottomrule
\end{tabular}
} 
\vspace{-7.5pt}
\caption{Comparison of accuracy and calibration performance using the CLIP-ViT/B16 backbone within the TPT framework under a predefined hard prompt (``a photo of a''). The best and second-best results are highlighted in \textbf{bold} and \underline{underline}, respectively.}
\label{tab:tpt_FG}
\vspace{-6pt}
\end{table*}

As discussed in Section~\ref{sec:Analysis},
existing regularization strategies for TPT encourage prompts to converge toward flat minima.
While they can improve calibration, the associated geometric constraints often distort output features, leading to performance degradation.
Moreover, because these methods rely on a single-step update,
they are fundamentally limited to effectively explore flat regions of the loss landscape.
Simply increasing the number of update steps is not a viable solution, as it incurs additional computational costs during TTA and amplifies the risk of overconfidence.

Motivated by the correlation between the sharpness of the loss landscape and calibration, we consider an alternative perspective for improving calibration in TPT:
rather than seeking flat minima around predefined prompts during TTA, we initialize the prompts directly within a flatter region, such that calibration becomes inherently better.
Formally, we aim to initialize prompts $\theta$ that already exhibit the desired flatness properties prior to TTA.
To achieve this,
we propose a flatness loss, which encourages the learned prompts to lie in flatter regions of the loss landscape by penalizing variations in the outputs under small perturbations:
\begin{equation}
\mathcal{L}_{\mathrm{flat}} =
\operatorname{dist}_{\text{cos}}\!\big(
  f_{T}(C+\varepsilon_{1};\;\theta+\varepsilon_{2}),\;
  f_{T}(C;\,\theta)
\big),
\label{eq:flatness_loss}
\end{equation}
where $\operatorname{dist}_{\text{cos}}$ denotes the cosine distance, and $\varepsilon_{1}$ and $\varepsilon_{2}$ are small random perturbations.
Intuitively, enforcing output stability under such perturbations reduces the model's sensitivity to changes in $\theta$, which leads to lower sharpness for any differentiable loss function (see Appendix~\ref{sec:Proof_1} for a formal proof).

However, optimizing only $\mathcal{L}_{\mathrm{flat}}$ can distort the original text features of $\theta_{0}^\text{zs}$, thereby degrading zero-shot performance.
Because the EM loss is highly sensitive to initial prediction probabilities~\cite{Wang2024}, such distortions can lead to substantial performance drops after adaptation.
To address this issue, we introduce an alignment loss that encourages the output text features of the learned prompt to remain close to those of $\theta_{0}^\text{zs}$ to preserve the semantic structure of the original prompt:
\begin{equation}
\mathcal{L}_{\mathrm{align}} =
\operatorname{dist}_{\text{L2}}\!\big(
  f_{T}(C;\,\theta),\;
  f_{T}(C;\,\theta_{0}^{\text{zs}})
\big),
\label{eq:align_loss}
\end{equation}
where $\operatorname{dist}_{\text{L2}}$ denotes the L2 distance. 
We then combine the two losses with a scaling factor $\lambda$ to form the final objective:
\begin{equation}
\mathcal{L}_{\mathrm{FPP}} = \mathcal{L}_{\mathrm{align}} + \lambda\, \mathcal{L}_{\mathrm{flat}}.
\label{eq:ours_loss}
\end{equation}
Here, we set $\lambda = \gamma_1 + \frac{\gamma_2}{K}$, where $\gamma_1, \gamma_2 > 0$ are hyperparameters and $K$ denotes the number of classes. This formulation downweights the flatness loss for larger class sets, where maintaining alignment with the original text features becomes more challenging.
Notably, both losses depend only on the predefined prompt $\theta_{0}^\text{zs}$ and the class-name set $C$, enabling data-free pretraining of prompts that inherit the calibration advantages of flat minima. 
The resulting prompt can be used as an initialization for existing TPT-based methods, which we apply without any modification to their adaptation procedures.

%% file: sec/6_Experiments.tex
\section{Experiments}
\label{sec:Experiments}

\subsection{Experimental Setting}\noindent

\noindent\textbf{Implementation Details.}
We use the CLIP-ViT-B/16 architecture as the backbone.
During the pretrain stage, we adopt the AdamW optimizer with a cosine learning rate scheduler. 
The initial learning rate is set to 0.01, and pretraining is performed for 1K iterations.  
For $\lambda$, we use $\gamma_1 = 1.0$ and $\gamma_2 = 0.15$.
The perturbation terms $\varepsilon_1$ and $\varepsilon_2$ are drawn from zero-mean isotropic Gaussian distributions with variances of 0.02 and 0.005, respectively.
Ablation studies for hyperparameter settings are provided in the Appendix~\ref{sec:hyperparam}.
For TTA, we follow the TPT~\cite{Shu2022} setting adopted in the baseline works~\cite{Yoon2024, Sharifdeen2025} without modification, and follow O-TPT~\cite{Sharifdeen2025} for other unspecified details.
Following the baseline works~\cite{Yoon2024, Sharifdeen2025}, we evaluate calibration using ECE and SCE, with detailed definitions provided in the Appendix~\ref{sec:metric}.

\begin{table*}[t!]
\centering
\resizebox{0.95\textwidth}{!}{%
\begin{tabular}{l|c|cccccccccc|c}
\toprule
\textbf{Method} & \textbf{Metric} & \textbf{Air} & \textbf{Calt} & \textbf{Car} & \textbf{DTD} & \textbf{SAT} & \textbf{FLW} & \textbf{Food} & \textbf{Pets} & \textbf{SUN} & \textbf{UCF} & \textbf{Avg.} \\
\midrule
\multirow{2}{*}{CoOp+TPT}        & Acc. &  20.0 & 94.0 & 65.6 & 44.5 & 40.6 & 68.7 & 83.8 & 89.1 & 65.6 & 67.2 & 63.91 \\
                                 & ECE & 29.6 & 3.65 & 6.63 & 34.8 & 31.3 & 19.9 & 9.66 & 7.40 & 20.8 & 19.9 & 18.36 \\
\midrule
\multirow{2}{*}{CoOp+C-TPT}     & Acc. &  19.2 & 93.9 & 63.1 & 45.0 & 40.7 & 69.0 & 83.7 & 89.3 & 65.1 & 66.6 & \underline{63.56} \\
                                & ECE & 21.5 & 1.66 & 2.45 & 21.0 & 13.2 & 10.2 & 4.49 & 2.12 & 11.8 & 12.0 & 10.04 \\
\midrule
\multirow{2}{*}{CoOp+O-TPT}     & Acc. & 18.69 & 93.71 & 64.12 & 45.45 & 40.17 & 68.57 & 83.55 & 89.07 & 64.01 & 65.64 & 63.29 \\
                                & ECE & 16.82 & 0.92 & 2.85 & 16.02 & 13.76 & 6.81 & 3.59 & 1.92 & 7.23 & 9.16 & \underline{7.91} \\
\midrule
\cellcolor{gray!20}      
        & \cellcolor{gray!20}Acc. &  \cellcolor{gray!20}21.51 & \cellcolor{gray!20}93.96 & \cellcolor{gray!20}63.74 & \cellcolor{gray!20}44.80 & \cellcolor{gray!20}48.72 & \cellcolor{gray!20}68.57 & \cellcolor{gray!20}84.31 & \cellcolor{gray!20}88.42 & \cellcolor{gray!20}66.44 & \cellcolor{gray!20}67.30 & \cellcolor{gray!20}\textbf{64.78} \\
\multirow{-2}{*}{\cellcolor{gray!20}CoOp+FPP (Ours)}                             
        & \cellcolor{gray!20}ECE  & \cellcolor{gray!20}9.34 & \cellcolor{gray!20}3.49 & \cellcolor{gray!20}5.33 & \cellcolor{gray!20}11.79 & \cellcolor{gray!20}9.69 & \cellcolor{gray!20}6.00 & \cellcolor{gray!20}1.21 & \cellcolor{gray!20}2.00 & \cellcolor{gray!20}3.00 & \cellcolor{gray!20}4.88 & \cellcolor{gray!20}\textbf{5.67} \\
\midrule
\midrule
\multirow{2}{*}{MaPLe+TPT}        & Acc. &  24.36 & 94.42 & 66.50 & 50.05 & 47.32 & 70.72 & 85.01 & 87.78 & 64.87 & 66.48 & \underline{65.75} \\
                                 & ECE & 10.58 & 2.38 & 4.14 & 11.80 & 9.42 & 11.63 & 1.78 & 1.79 & 8.47 & 7.41 & 6.94 \\
\midrule
\multirow{2}{*}{MaPLe+O-TPT}     & Acc. & 24.00 & 92.29 & 65.38 & 49.11 & 44.58 & 71.53 & 84.35 & 89.97 & 63.49 & 65.82 & 65.05 \\
                                & ECE & 6.41 & 3.49 & 3.61 & 4.90 & 7.92 & 4.35 & 1.49 & 3.97 & 2.78 & 2.22 & \underline{4.11} \\
\midrule
\cellcolor{gray!20}      
        & \cellcolor{gray!20}Acc. &  \cellcolor{gray!20}24.06 & \cellcolor{gray!20}93.47 & \cellcolor{gray!20}66.27 & \cellcolor{gray!20}48.94 & \cellcolor{gray!20}52.19 & \cellcolor{gray!20}69.35 & \cellcolor{gray!20}85.17 & \cellcolor{gray!20}91.55 & \cellcolor{gray!20}68.29 & \cellcolor{gray!20}70.84 & \cellcolor{gray!20}\textbf{67.01} \\
\multirow{-2}{*}{\cellcolor{gray!20}MaPLe+FPP (Ours)}                            
        & \cellcolor{gray!20}ECE  & \cellcolor{gray!20}3.98 & \cellcolor{gray!20}2.83 & \cellcolor{gray!20}2.18 & \cellcolor{gray!20}11.58 & \cellcolor{gray!20}1.91 & \cellcolor{gray!20}4.03 & \cellcolor{gray!20}3.53 & \cellcolor{gray!20}5.03 & \cellcolor{gray!20}1.74 & \cellcolor{gray!20}3.93 & \cellcolor{gray!20}\textbf{4.07} \\
\bottomrule
\end{tabular}
} 
\vspace{-7.5pt}
\caption{Comparison of accuracy and calibration error using the CLIP-ViT/B16 backbone within the TPT framework under predefined prompts trained from CoOp and MAPLE. The best and second-best results are highlighted in \textbf{bold} and \underline{underline}, respectively.}
\label{tab:tpt_init}
\vspace{-4.0pt}
\end{table*}

\begin{table*}[t!]
\centering
\resizebox{0.95\textwidth}{!}{%
\begin{tabular}{l|c|ccccccccc|c}
\toprule
\textbf{Method} & \textbf{Metric} & \textbf{Air} & \textbf{Calt} & \textbf{Car} & \textbf{DTD} & \textbf{SAT} & \textbf{FLW} & \textbf{Food} & \textbf{Pets} & \textbf{UCF} & \textbf{Avg.} \\
\midrule
\multirow{1}{*}{DynaPrompt (Paper)}        & Acc. &  24.33 & 94.32 & 67.65 & 47.96 & 42.28 & 69.95 & 85.42 & 88.28 & 68.72 & 65.43 \\
\midrule
\midrule
\multirow{3}{*}{DynaPrompt (Replication)}        & Acc. & 22.68 & 94.16 & 66.88 & 47.87 & 35.91 & 69.67 & 84.92 & 87.71 & 68.17 & \underline{64.22} \\
                                 & ECE & 18.67 & 3.05 & 5.08 & 23.19 & 33.76 & 12.73 & 6.75 & 5.62 & 13.72 & 13.62 \\
                                 & SCE & 0.73 & 0.15 & 0.21 & 1.53 & 8.53 & 0.52 & 0.18 & 0.57 & 0.49 & 1.43 \\
                                 
\midrule
\multirow{3}{*}{DynaPrompt+C-TPT}     & Acc. & 23.34 & 93.83 & 66.04 & 46.99 & 36.57 & 70.08 & 83.58 & 88.61 & 65.74 & 63.86 \\
                                & ECE &  10.21 & 2.51 & 2.17 & 15.87 & 16.94 & 4.35 & 2.21 & 2.43 & 5.33 & 6.89 \\
                                & SCE &  0.63 & 0.19 & 0.22 & 1.43 & 6.99 & 0.53 & 0.19 & 0.57 & 0.49 & \underline{1.25} \\
\midrule
\multirow{3}{*}{DynaPrompt+O-TPT}     & Acc. & 22.41 & 93.67 & 65.58 & 45.80 & 36.32 & 68.74 & 83.39 & 88.55 & 65.69 & 63.35 \\
                                & ECE & 9.60 & 2.74 & 2.13 & 14.31 & 17.73 & 3.60 & 2.03 & 2.78 & 3.43 & \underline{6.48} \\
                                & SCE & 0.61 & 0.19 & 0.23 & 1.43 & 7.04 & 0.55 & 0.20 & 0.58 & 0.51 & 1.26 \\
\midrule
\cellcolor{gray!20}      
                                & \cellcolor{gray!20}Acc. & \cellcolor{gray!20}24.96 & \cellcolor{gray!20}92.33 & \cellcolor{gray!20}66.26 & \cellcolor{gray!20}46.10 & \cellcolor{gray!20}49.09 & \cellcolor{gray!20}69.63 & \cellcolor{gray!20}84.51 & \cellcolor{gray!20}88.36 & \cellcolor{gray!20}66.09 & \cellcolor{gray!20}\textbf{65.26} \\
\cellcolor{gray!20}                                
                                & \cellcolor{gray!20}ECE  & \cellcolor{gray!20}5.28 & \cellcolor{gray!20}5.72 & \cellcolor{gray!20}3.70 & \cellcolor{gray!20}5.27 & \cellcolor{gray!20}8.49 & \cellcolor{gray!20}2.44 & \cellcolor{gray!20}2.94 & \cellcolor{gray!20}4.96 & \cellcolor{gray!20}2.20 & \cellcolor{gray!20}\textbf{4.56} \\
\multirow{-3}{*}{\cellcolor{gray!20}DynaPrompt+FPP (Ours)}
                                & \cellcolor{gray!20}SCE & \cellcolor{gray!20}0.56 & \cellcolor{gray!20}0.28 & \cellcolor{gray!20}0.23 & \cellcolor{gray!20}1.33 & \cellcolor{gray!20}5.72 & \cellcolor{gray!20}0.59 & \cellcolor{gray!20}0.21 & \cellcolor{gray!20}0.70 & \cellcolor{gray!20}0.53 & \cellcolor{gray!20}\textbf{1.13} \\
\bottomrule
\end{tabular}
} 
\vspace{-7.5pt}
\caption{Comparison of accuracy and calibration error using the CLIP-ViT/B16 backbone within the DynaPrompt framework under a predefined hard prompt (``a photo of a''). The best and second-best results are highlighted in \textbf{bold} and \underline{underline}, respectively.}
\label{tab:dynaprompt_FG}
\vspace{-14.0pt}
\end{table*}

\subsection{Main Results}
\noindent\textbf{Fine-Grained Classification.}
Tab.~\ref{tab:tpt_FG} presents the results for the fine-grained classification task, where the predefined prompt $\theta_{0}^{\text{zs}}$ is set as a hard textual template ``a photo of a.'' with class names provided for each dataset.
In terms of calibration errors, our method achieves state-of-the-art (SOTA) performance in both ECE and SCE, with a particularly large improvement in SCE. 
Moreover, our approach also achieves SOTA performance in accuracy, even surpassing TPT. 
This suggests that the sharpness of the loss landscape can also influence the post-adaptation accuracy.
Notably, while C-TPT and O-TPT exhibit a trade-off between accuracy and calibration, our method is the first to achieve SOTA performance in both metrics simultaneously.

\vspace{3pt}\noindent\textbf{Different Predefined Prompts.}
Tab.~\ref{tab:tpt_init} presents the results under the same fine-grained classification datasets as in Tab.~\ref{tab:tpt_FG}, but with the predefined prompt $\theta_{0}^{\text{zs}}$ obtained from the supervised-trained prompt embeddings. 
Following the baseline~\cite{Sharifdeen2025}, we employ the officially released checkpoints of CoOp~\cite{Zhou2022b} and MaPLe~\cite{Khattak2023a}, and use them to train a new initial prompt $\theta$ according to Eq.~\eqref{eq:ours_loss}. 
Our approach achieves SOTA performance in both accuracy and calibration, showing even larger improvements compared to the hard prompt setting in Tab.~\ref{tab:tpt_FG}. 
These results indicate that our method can leverage prior knowledge more effectively, suggesting its strong compatibility with supervised approaches.

\vspace{3pt}\noindent\textbf{Different TPT Framework.}
Tab.~\ref{tab:dynaprompt_FG} presents the results obtained using the Dynaprompt~\cite{Xiao2025} framework, which accumulates updates during TTA instead of resetting the prompt for each sample as in TPT.
Specifically, Dynaprompt maintains multiple prompts, adaptively selects suitable ones for each sample, and accumulates updates during TTA.
Since the selected prompts are optimized using the same EM loss as in TPT, we simply add the regularization terms from the baselines~\cite{Yoon2024, Sharifdeen2025} to apply them within this new framework.
The predefined prompt $\theta_{0}^\text{zs}$ is set to ``a photo of a'' following the original setup, with class names provided for each dataset.
We adjust the prompt buffer size to 12 and reproduce all experiments using their official codebase. 
Our method still achieves SOTA performance across all evaluation metrics, showing strong generalizability in different learning paradigms.

\vspace{3pt}\noindent\textbf{Natural Distribution Shifts.}
Tab.~\ref{tab:tpt_OOD} presents results on natural distribution shifts~\cite{Zhou2022b}, evaluating out-of-distribution (OOD) performance using widely adopted ImageNet variant datasets, where details provided in Appendix.
In this setting, we fix $\lambda$ as 1.25 across all datasets.
Under this configuration, our method consistently achieves notable improvements in both ECE and SCE, with only a minor accuracy reduction of 0.42\% compared to TPT.
In contrast, O-TPT, which exhibits the best calibration among prior works, experiences an accuracy drop of more than 2\%.
These findings highlight that our method maintains strong effectiveness under OOD scenarios.

\begin{table}[t!]
\centering
\resizebox{0.94\columnwidth}{!}{%
\begin{tabular}{l|c|ccccc|c}
\toprule
\textbf{Method} & \textbf{Metric} & \textbf{I} & \textbf{I-A} & \textbf{I-V2} & \textbf{I-R} & \textbf{I-S} & \textbf{OOD Avg.} \\
\midrule
\multirow{3}{*}{CLIP~\cite{Radford2021}} & Acc. &  66.7 & 47.8 & 60.8 & 74.0 & 46.1 & 57.18 \\
                                              & ECE  &  2.12 & 8.61 & 3.01 & 3.58 & 4.95 & 5.04 \\
                                              & SCE  &  0.04 & 0.30 & 0.06 & 0.18 & 0.06 & 0.15 \\
\midrule    
\multirow{3}{*}{TPT~\cite{Shu2022}}           & Acc. &  69.0 & 52.6 & 63.0 & 76.7 & 47.5 & \textbf{59.95} \\
                                              & ECE  &  10.6 & 16.4 & 11.1 & 4.36 & 16.1 & 11.99 \\
                                              & SCE  & 0.04 & 0.29 & 0.06 & 0.14 & 0.06 & 0.14 \\
\midrule         
\multirow{3}{*}{C-TPT~\cite{Yoon2024}}        & Acc. &  68.5 & 51.6 & 62.7 & 76.0 & 47.9 & 59.55 \\
                                              & ECE  &  3.15 & 8.16 & 6.23 & 1.54 & 7.35 & 5.82 \\
                                              & SCE  & 0.04 & 0.29 & 0.06 & 0.16 & 0.06 & 0.14 \\
\midrule   
\multirow{3}{*}{O-TPT~\cite{Sharifdeen2025}}  & Acc. &  67.3 & 49.9 & 61.7 & 72.6 & 47.1 & 57.82 \\
                                              & ECE  &  1.97 & 7.22 & 3.97 & 1.46 & 6.87 & 4.88 \\
                                              & SCE  & 0.04 & 0.29 & 0.06 & 0.17 & 0.06 & 0.15 \\
\midrule   
\cellcolor{gray!20}                         & \cellcolor{gray!20}Acc. &  \cellcolor{gray!20}67.8 & \cellcolor{gray!20}52.3 & \cellcolor{gray!20}61.9 & \cellcolor{gray!20}76.7 & \cellcolor{gray!20}47.2 & \cellcolor{gray!20}59.53 \\
\cellcolor{gray!20}                                             
                                            & \cellcolor{gray!20}ECE  &  \cellcolor{gray!20}2.97 & \cellcolor{gray!20}5.38 & \cellcolor{gray!20}3.45 & \cellcolor{gray!20}7.07 & \cellcolor{gray!20}2.62 & \cellcolor{gray!20}\textbf{4.63} \\
\multirow{-3}{*}{\cellcolor{gray!20}FPP (Ours)}
                                            & \cellcolor{gray!20}SCE & \cellcolor{gray!20}0.04 & \cellcolor{gray!20}0.27 & \cellcolor{gray!20}0.06 & \cellcolor{gray!20}0.16 & \cellcolor{gray!20}0.06 & \cellcolor{gray!20}\textbf{0.12} \\
\bottomrule
\end{tabular}
} 
\vspace{-8.0Pt}
\caption{Comparison of accuracy and calibration error in natural distribution shifts datasets.}
\label{tab:tpt_OOD}
\vspace{-10.0pt}
\end{table}

\begin{table}[t!]
  \centering
  \resizebox{0.94\columnwidth}{!}{%
    \begin{tabular}{%
      >{\centering\arraybackslash}p{0.18\linewidth}
      >{\centering\arraybackslash}p{0.09\linewidth}
      >{\centering\arraybackslash}p{0.09\linewidth}|
      >{\centering\arraybackslash}p{0.18\linewidth}|
      >{\centering\arraybackslash}p{0.1\linewidth}
      >{\centering\arraybackslash}p{0.1\linewidth}
      >{\centering\arraybackslash}p{0.1\linewidth}}
      \toprule
      \multirow[c]{2}{*}{\textbf{Align Loss}} &
      \multicolumn{2}{c|}{\textbf{Flat Loss}} &
      \multirow[c]{2}{*}{\textbf{Zero-Shot}} &
      \multirow[c]{2}{*}{\textbf{Acc.}} &
      \multirow[c]{2}{*}{\textbf{ECE}} &
      \multirow[c]{2}{*}{\textbf{SCE}} \\
      \cmidrule(lr){2-3}
       & $\varepsilon_{1}$ & $\varepsilon_{2}$ & & & \\
      \midrule
      -- & -- & -- & 63.41 & 64.62 & 11.67 & 1.15 \\
      -- & \checkmark & \checkmark & 4.65 & 4.16 & -- & -- \\
      \checkmark & -- & -- & 63.49 & 64.40 & 7.05 & 1.09 \\      \midrule
      \checkmark & -- & \checkmark & 63.92 & 64.77 & 5.86 & 1.03 \\
      \checkmark & \checkmark & -- & 64.15 & 64.89 & 4.64 & 1.02 \\
      \midrule
      \cellcolor{gray!20}\checkmark & \cellcolor{gray!20}\checkmark & \cellcolor{gray!20}\checkmark & \cellcolor{gray!20}\textbf{64.35} & \cellcolor{gray!20}\textbf{65.37} & \cellcolor{gray!20}\textbf{4.13} & \cellcolor{gray!20}\textbf{0.96} \\
      \bottomrule
    \end{tabular}
  }
  \vspace{-8.0pt}
  \caption{Ablation study under the fine-grained classification setting, evaluating the contribution of each component.}
  \label{tab:ablation}
  \vspace{-10.0pt}
\end{table}

\subsection{Experimental Analysis}\noindent

\noindent\textbf{Ablation Study.}
Tab.~\ref{tab:ablation} summarizes the ablation results across fine-grained classification datasets using TPT, designed to assess the effectiveness of the two proposed losses, the $\mathcal{L}_{\mathrm{align}}$ and the $\mathcal{L}_{\mathrm{flat}}$, in the pretraining stage.
The first block reports the default setting, where TPT is directly applied to the original initial prompt $\theta_{0}^\text{zs}$ without any pretraining.
The second block shows that employing the the flatness loss $\mathcal{L}_{\mathrm{flat}}$ without the alignment loss $\mathcal{L}_{\mathrm{align}}$ leads to a collapse in the zero-shot accuracy of the learned prompt $\theta$, making adaptation with the EM loss infeasible.
Conversely, applying only the $\mathcal{L}_{\mathrm{align}}$ preserves baseline zero-shot performance and achieves competitive results after TPT adaptation, but it does not ensure proper calibration.
The fourth and fifth blocks ablate the effect of each perturbation component within the $\mathcal{L}_{\mathrm{flat}}$, showing that both contribute to improvements in accuracy and calibration.
Finally, the last row confirms that combining all components yields the best overall performance, indicating that each component contributes synergistically to performance improvement.

\vspace{3pt}\noindent\textbf{Class Name Dependency.}
In many TTA methods, task-specific information such as the set of class names $C$ is known in advance and often utilized to design task-specific configurations~\cite{Zhu2024, Zhang2024, Karmanov2024}.
However, this assumption may not hold in scenarios where such information is unavailable prior to test-time.
To address this issue, we conduct experiments summarized in Tab~\ref{tab:class_dependency}, where pretraining is performed after modifying the original class names in two ways: (i) applying ImageNet class names, and (ii) replacing the original class name embeddings with Gaussian noise.
Notably, our method consistently achieves SOTA performance in both cases, surpassing the baseline results reported in Tab.~\ref{tab:tpt_FG}.
This finding aligns with observations from~\cite{Zheng2023}, which indicate that text semantics are relatively easy to sample and that even random text embeddings can effectively preserve the underlying feature space.
Overall, these results demonstrate that our approach is largely independent of explicit class-name supervision and exhibits strong generalization, even when class names are partially or entirely absent.

\begingroup
\setlength{\tabcolsep}{4pt}       
\renewcommand{\arraystretch}{0.90}
\begin{table}[t]
  \centering
  \resizebox{0.8\columnwidth}{!}{
    \begin{tabular}{c|c|c|c}
      \toprule
      \textbf{Class Source} & \textbf{Metric} & \textbf{Fine-Grained Avg.} & \textbf{$\Delta$} \\
      \midrule
      \multirow[c]{3}{*}{ImageNet-1K}
        & Acc. & 64.97 & -0.40 \\
        & ECE  & 4.40 & +0.27 \\
        & SCE  & 0.97 & +0.01 \\
      \midrule
      \multirow[c]{3}{*}{\shortstack{Gaussian\\Noise}}
        & Acc. & 65.28 & -0.09 \\
        & ECE  & 4.25 & +0.12 \\
        & SCE  & 1.03 & +0.07 \\
      \bottomrule
    \end{tabular}
  }
  \vspace{-7.0pt}
  \caption{Evaluation of robustness to class name variation, reported as the average performance across fine-grained classification datasets. $\Delta$ denotes the performance difference relative to the default setting, which uses original class names during pretraining.}
  \label{tab:class_dependency}
  \vspace{-14.0pt}
\end{table}
\endgroup

\begin{figure*}[t!]
\centering
\begin{minipage}[t]{0.49\textwidth}
\centering
\includegraphics[width=\linewidth]{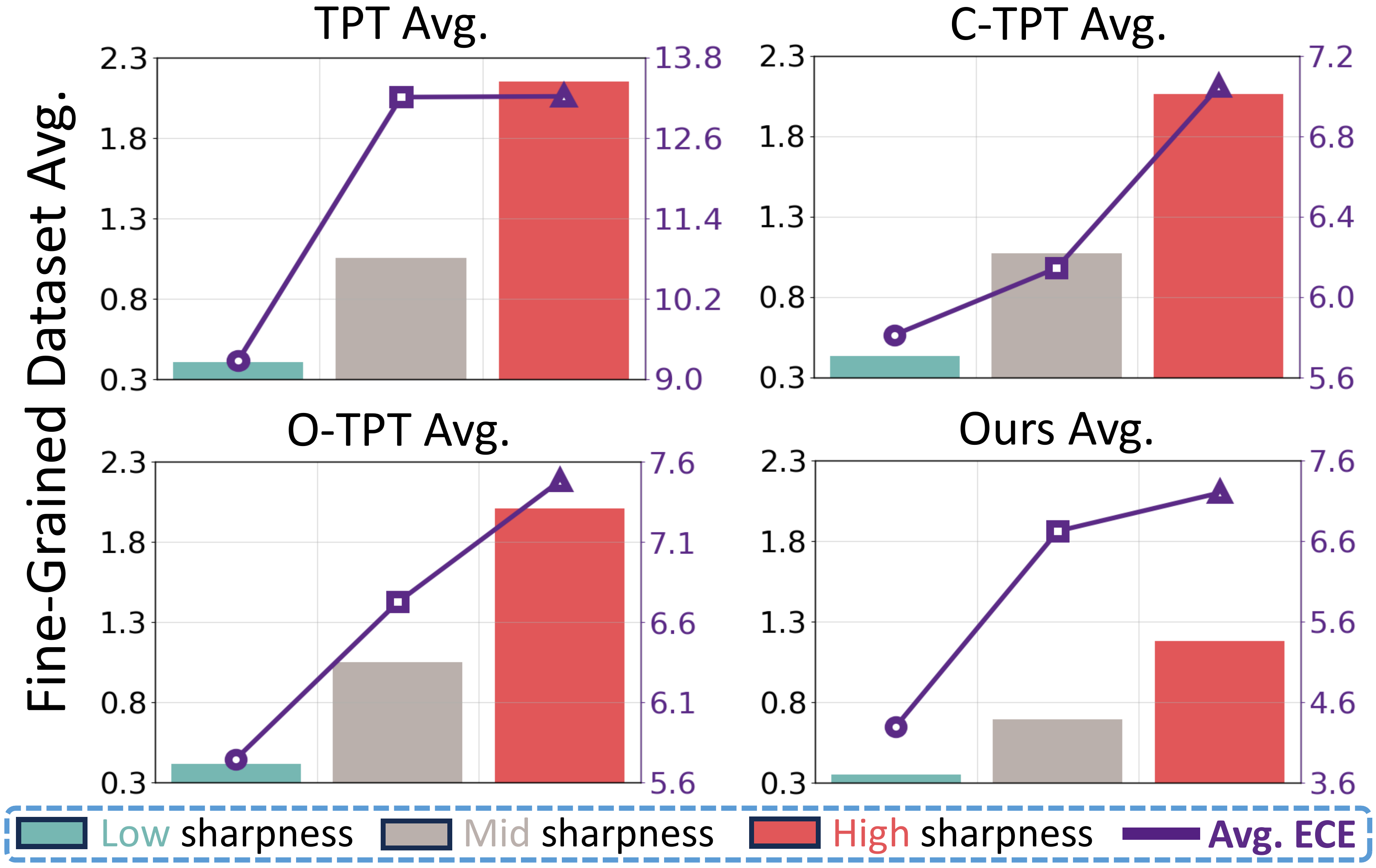}
\vspace{-20.0pt}
\captionof{figure}{Relationship between sharpness of the loss landscape and calibration error. Across several datasets and methods, flatter minima consistently correspond to lower calibration errors.}
\label{fig:flatness2performance}
\end{minipage}
\hfill
\begin{minipage}[t]{0.49\textwidth}
\centering
\includegraphics[width=\linewidth]{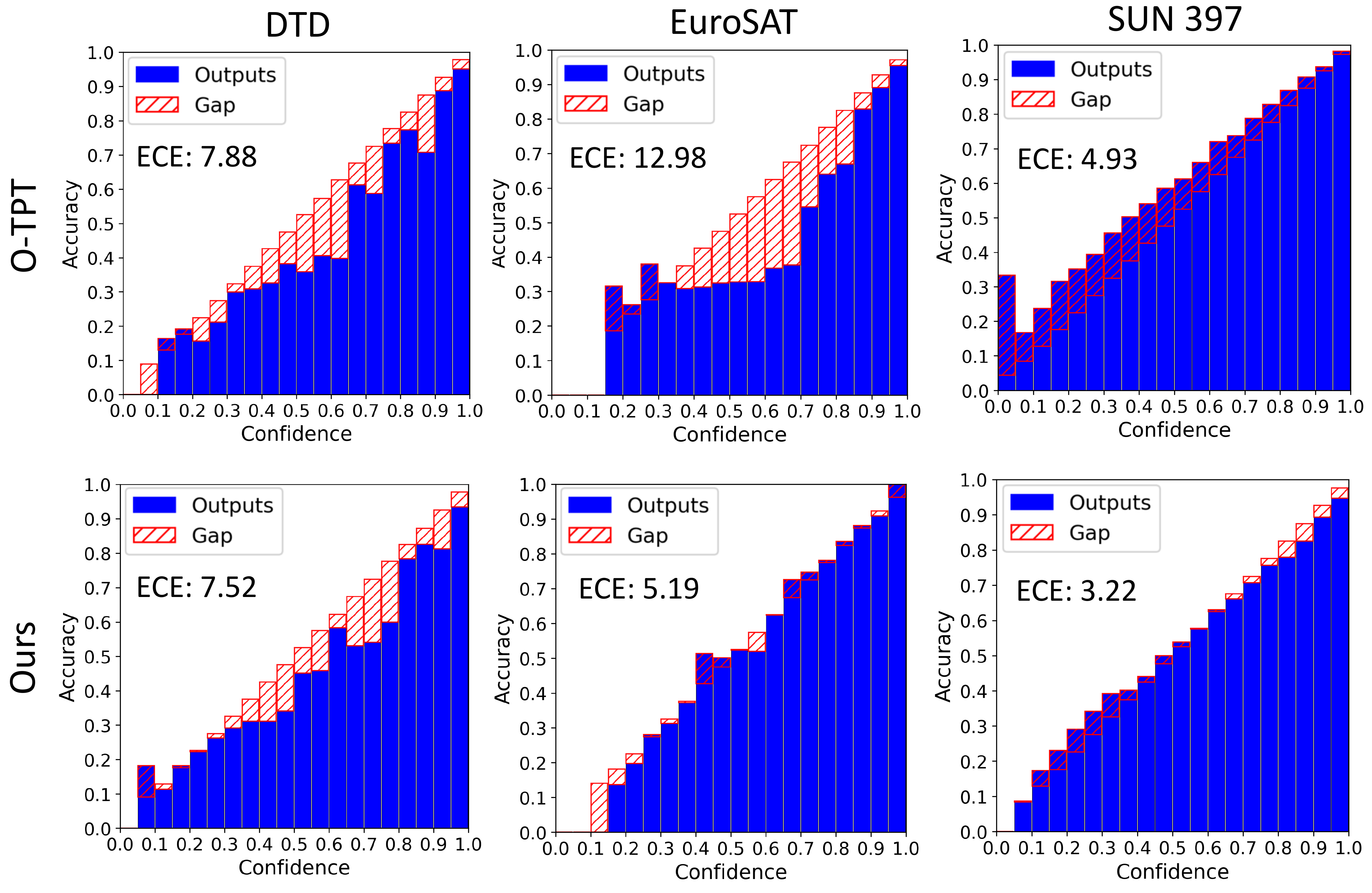}
\vspace{-20.0pt}
\captionof{figure}{Reliability diagrams showing underconfidence and overconfidence. Our method demonstrates the ability to produce appropriate confidence values as needed.}
\label{fig:reliability}
\end{minipage}
\vspace{-8.0pt}
\end{figure*}

\vspace{3pt}\noindent\textbf{Flat Minima and Calibration.}
To further investigate the relationship between flat minima and calibration, we conduct the experiments shown in Fig.~\ref{fig:flatness2performance}.
For each test sample, we first optimize an individual prompt and then quantify the sharpness of the corresponding loss landscape.
The resulting prompts are sorted in ascending order of sharpness and divided equally into three groups, representing low, medium, and high sharpness levels.
Within each dataset, we compute the ECE and mean sharpness for each group, and then average these quantities across all fine-grained datasets to visualize their relationship.
This procedure follows the baseline setup in~\cite{Sharifdeen2025}, ensuring that each group contains an equal and sufficient number of samples for statistically reliable calibration estimates.
Across different calibration methods, our findings consistently show that prompts converging to flatter minima exhibit lower calibration errors.
These results suggest that, among the various local minima of the EM loss, those corresponding to flatter minima yield more reliable predictive probabilities.

\begin{table*}[t!]
\centering
\begin{minipage}[t]{0.33\textwidth}
\centering
\resizebox{\linewidth}{!}{
  \begin{tabular}{c|ccc|c}
    \toprule
    Method & Acc. & ECE & SCE & Sharp. \\
    \midrule
    TPT  & 64.62 & 11.67 & 1.15 & 1.23 \\
    TPT+ASAM~\cite{Kwon2021} & 64.18 & 9.78 & 1.12 & 1.12 \\
    TPT+SAM~\cite{Foret2021}  & 61.59 & 4.15 & 1.12 & 0.78 \\
    \midrule
    \cellcolor{gray!20}FPP (Ours) & \cellcolor{gray!20}\textbf{65.37} & \cellcolor{gray!20}\textbf{4.13} & \cellcolor{gray!20}\textbf{0.96} & \cellcolor{gray!20}\textbf{0.52} \\
    \bottomrule
  \end{tabular}
}
\vspace{-5.0pt}
\captionof{table}{Evaluation of SAM-based methods under the TPT setting, showing reduced calibration error but degraded accuracy.} 
\label{tab:sam}
\end{minipage}
\hfill
\begin{minipage}[t]{0.345\textwidth}
\centering
\resizebox{\linewidth}{!}{
  \begin{tabular}{c|ccc|c}
    \toprule
    Method & Acc. & ECE & SCE & Sharp. \\
    \midrule
    $L_\text{ent} \rightarrow L_\text{ent}$ (2-step)  & 64.67 & 18.90 & 1.27 & 1.31 \\
    $L_\text{ent} \rightarrow L_\text{reg}$ (2-step) & 64.14 & 12.60 & 1.17 & 1.26 \\
    $L_\text{reg} \rightarrow L_\text{ent}$ (2-step)  & 64.29 & 11.78 & 1.19 & 1.22 \\
    \midrule
     $L_\text{ent} + L_\text{reg}$ (O-TPT)  & 64.12 & 4.46 & 1.15 & 1.16 \\
    \bottomrule 
  \end{tabular}
}
\vspace{-5.0pt}
\captionof{table}{Evaluation of isolated effect of regularization loss in O-TPT, showing limited calibration benefit without SAM mechanism.}
\label{tab:isolated_effect}
\end{minipage}
\hfill
\begin{minipage}[t]{0.29\textwidth}
\centering
\resizebox{\linewidth}{!}{
  \begin{tabular}{l|l|cc}
    \toprule
    Method & Dataset & Pretrain & TTA \\
    \midrule
    \multirow{2}{*}{O-TPT} & EuroSAT & --  & 9m 55s \\
                           & SUN397  & --  & 81m 27s\\
    \midrule
      \cellcolor{gray!20} & \cellcolor{gray!20}EuroSAT & \cellcolor{gray!20}0m 18s & \cellcolor{gray!20}\textbf{9m 11s} \\
      \multirow{-2}{*}{\cellcolor{gray!20} FPP (Ours)} & \cellcolor{gray!20}SUN397  & \cellcolor{gray!20}10m 21s  & \cellcolor{gray!20}\textbf{76m 50s} \\
    \bottomrule
  \end{tabular}
}
\vspace{-5.0pt}
\captionof{table}{Comparison of computational costs in terms of execution time between our method and O-TPT.}
\label{tab:computational_costs}
\end{minipage}
\vspace{-8.0pt}

\end{table*}

\vspace{3pt}\noindent\textbf{Underconfidence and Overconfidence.}
To examine whether our approach can address both overconfidence and underconfidence, we present reliability diagrams in Fig.~\ref{fig:reliability}.
As illustrated, O-TPT exhibits overconfidence on the texture dataset (DTD) and the remote-sensing dataset (EuroSAT), while showing underconfidence on the natural-domain dataset (SUN397).
In contrast, our method effectively mitigates these issues across all cases, showing consistent improvements over both types of calibration error.

\vspace{3pt}\noindent\textbf{TPT with SAM Objective.}
In Section~\ref{sec:Analysis}, we show that the optimization formulation of regularized TPT~\cite{Sharifdeen2025, Yoon2024} is inherently aligned with the core principle underlying the SAM mechanism.
The key distinction, however, is that regularized TPT operates with a fixed perturbed point, defined as $\theta_{0}^\text{zs}=\theta_{0}^\text{reg}+\varepsilon_\text{reg}$, which we argue is key to enabling SAM in TPT.
When SAM-based methods~\cite{Foret2021, Kwon2021} are directly applied to TPT, the algorithm simply seeks a perturbation direction that increases the EM loss, without considering the prediction probabilities associated with the perturbed prompts.
Such perturbations often distort text features, causing incorrect predictions after adaptation due to the high sensitivity of the EM loss to initial accuracy~\cite{Wang2024}.
As shown in Table~\ref{tab:sam}, although these approaches successfully reduce both sharpness and calibration error, they also lead to notable accuracy degradation.
One could attempt to mitigate this issue by filtering out noisy test samples to avoid perturbations that destabilize initial predictions~\cite{Niu2023, Gong2023}. 
However, this strategy is infeasible in TPT, where sample-specific adaptation is essential. 
In contrast, our method enables the prompt to converge to flat minima without relying on any form of perturbation.

\vspace{3pt}\noindent\textbf{Regularization Loss Without the SAM Effect.}
As noted earlier, when the regularization loss is combined with the EM loss, it follows the same optimization principle as SAM, and we argued that the resulting flat minima are the reason of improved calibration.
At this point, one might question whether the regularization loss itself has an intrinsic influence on calibration, independent of the flat minima.
To address this concern, we conduct experiments in which the two losses are optimized sequentially rather than jointly.
This update scheme no longer aligns with the SAM formulation and, as shown in Tab.~\ref{tab:isolated_effect}, it fails to produce any meaningful reduction in sharpness.
Correspondingly, we observe no improvement in calibration, and the geometric constraint imposed by the regularization term even degrades accuracy.
This finding highlights that it is the sharpness of the loss landscape---rather than the regularization term itself---that plays a direct role in achieving effective calibration.

\vspace{3pt}\noindent\textbf{Computational Costs.}
Tab.~\ref{tab:computational_costs} summarizes the computation times for both the pretraining and TTA stages.
Note that the proposed pretraining method does not rely on any test samples and is completely decoupled from the inference process.
Consequently, our method only affects the pretraining stage, maintaining the same inference time as the original TPT, while the overall pretraining time remains negligible compared to the inference time.
In contrast, O-TPT, in its practical implementation, incorporates a Householder transformation~\cite{Kaufman1987} within the regularization loss, introducing a $\mathcal{O}(|C|^3)$ computational complexity and causing a direct increase in inference latency.

%% file: sec/7_Conclusion.tex
\section{Conclusion}
\label{sec:Conclusion}
In this paper, we reveal that prompts converging to flat minima consistently achieve superior calibration performance compared to those trapped in sharper minima. 
We further show that existing regularization methods implicitly function as mechanisms that guide prompts toward flat minima of the EM loss.
Building on these insights, we propose FPP, a pretraining framework that positions prompts in flatter regions before adaptation, facilitating more effective convergence to flat minima.
Notably, simply replacing the initial prompt in existing methods with our pretrained ones yields SOTA calibration and accuracy.

%% file: sec/8_Acknowledgment.tex
\section{Acknowledgments}
This work was partially supported by the National Research Foundation of Korea (NRF) grant funded by the Ministry of Science and ICT (MSIT) of the Korean government (RS-2024-00341749), and Institute of Information \& Communications Technology Planning \& Evaluation (IITP) grant funded by MSIT (RS-2022-II220124, RS-2023-00259934, RS-2025-02283048).

%% file: sec/X_suppl.tex
\clearpage
\setcounter{page}{1}
\maketitlesupplementary

\appendix
\numberwithin{table}{section}
\numberwithin{figure}{section}
\numberwithin{equation}{section}

\section{Proof of Theorem 1}
\label{sec:Theory_1}

In the main paper, Theorem~\ref{thm:reg_ent_corr} states that, for either regularizer associated with Eq.~\eqref{eq:ctpt_detail} or Eq.~\eqref{eq:otpt_detail},
\begin{equation}
\begin{aligned}
\mathcal H(T)
&\coloneq \mathbb E_{v\sim \mathrm{Unif}(\mathbb S^{D-1})}
\bigl[L_{\mathrm{ent}}(T,v)\bigr] \\
&= \alpha L_{\mathrm{reg}}(T)
+ \beta
+ O(D^{-3/2}),
\end{aligned}
\end{equation}
with \(\alpha>0\). We now make this statement precise and prove it.

Recall that \(T \in \mathbb{R}^{K \times D}\) denotes the text feature matrix, and \(t_i \in \mathbb{R}^{D}\) denotes its \(i\)-th row, where \(\|t_i\|_2 = 1\). This assumption is imposed by the CLIP model, which normalizes feature embeddings and computes cosine similarity. 
To facilitate the analysis, we introduce equivalent surrogate losses that share the same optimal states as the original regularizers:
\begin{equation}
L_{\mathrm{disp}}(T)\coloneq-\|PT\|_F^2,
\qquad
P\coloneq I-\frac{1}{K}11^\top,
\end{equation}
for the regularizer in Eq.~\eqref{eq:ctpt_detail}, and
\begin{equation}
L_{\mathrm{orth}}(T)
\coloneq
\sum_{1\le i<j\le K} t_i^\top t_j,
\end{equation}
for the regularizer in Eq.~\eqref{eq:otpt_detail}. 
We will show that both losses depend only on the same term, and that the expected EM loss is also determined by that term.

Let \(\mu(T)=\frac{1}{K}\sum_{i=1}^K t_i\) denote the mean text feature. We then define the following quantity:
\begin{equation}
\begin{aligned}
S(T)
&\coloneq \|PT\|_F^2 \\
&= \sum_{i=1}^K \|t_i-\mu(T)\|_2^2.
\end{aligned}
\end{equation}

We first relate \(L_{\mathrm{disp}}(T)\) and
\(L_{\mathrm{orth}}(T)\) to \(S(T)\).
Since \(\sum_{i=1}^K (t_i-\mu(T))=0\) and \(\|t_i\|_2=1\), we have
\begin{equation}
S(T)=K-K\|\mu(T)\|_2^2.
\end{equation}
Moreover, since each \(t_i\) has unit norm, \(\|\mu(T)\|_2^2\) is determined by the pairwise inner products:
\begin{equation}
\begin{aligned}
\|\mu(T)\|_2^2
&= \frac{1}{K^2}\Bigl\|\sum_{i=1}^K t_i\Bigr\|_2^2 \\
&= \frac{1}{K^2}\left(
K + 2\sum_{1\le i<j\le K} t_i^\top t_j
\right).
\end{aligned}
\end{equation}
Substituting this into the expression for \(S(T)\), we obtain
\begin{equation}
S(T)=K-1-\frac{2}{K}\sum_{1\le i<j\le K} t_i^\top t_j.
\end{equation}
Therefore,
\begin{equation}
L_{\mathrm{disp}}(T)=-S(T),
\qquad
L_{\mathrm{orth}}(T)=\frac{K}{2}\bigl(K-1-S(T)\bigr).
\end{equation}
Thus, both losses depend only on \(S(T)\).

We next relate \(S(T)\) to the expected EM loss.
For an image feature \(v\sim \mathrm{Unif}(\mathbb S^{D-1})\), let
\begin{equation}
s(T,v)\coloneq Tv
\end{equation}
denote the logit vector, and define
\begin{equation}
f(s)\coloneq H\!\bigl(\mathrm{softmax}(s)\bigr),
\end{equation}
where \(H\) denotes entropy.
Since softmax is invariant under adding the same scalar to all logits, we have
\begin{equation}
f(s+c1)=f(s)=f(Ps)
\qquad
\text{for all } c\in\mathbb R,
\end{equation}
where \(Ps=s-\frac{1}{K}(1^\top s)1\).
Thus, \(f(s)\) is fully determined by the centered logits \(Ps\).

We now expand \(f\) around the origin. Since \(\|t_i\|_2=\|v\|_2=1\), each logit satisfies \(|s_i(T,v)|=|t_i^\top v|\le 1\). Therefore, \(s(T,v)\in[-1,1]^K\), and the centered logits \(Ps(T,v)\) lie in the compact set \(\{Ps : s\in[-1,1]^K\}\).
As \(f\) is smooth, its third derivatives are uniformly bounded on this set. At \(s=0\), the softmax distribution is uniform, so
\begin{equation}
f(0)=\log K,
\qquad
\nabla f(0)=0,
\qquad
\nabla^2 f(0)=-\frac{1}{K}P.
\end{equation}
Hence, Taylor expansion around the origin gives
\begin{equation}
f(s)
=
\log K
-
\frac{1}{2K}\|Ps\|_2^2
+
R(s),
\end{equation}
where \(R(s)\) is the remainder term satisfying
\begin{equation}
|R(s)|\le M_K\|Ps\|_2^3
\end{equation}
for some constant \(M_K<\infty\) depending only on \(K\).

Substituting \(s(T,v)=Tv\) and taking expectation over
\(v\sim \mathrm{Unif}(\mathbb S^{D-1})\), we obtain
\begin{equation}
\begin{aligned}
\mathcal H(T)
&=
\mathbb E_v\bigl[L_{\mathrm{ent}}(T,v)\bigr] \\
&=
\log K
-
\frac{1}{2K}\mathbb E_v\|PTv\|_2^2
+
\mathbb E_v[R(Tv)].
\end{aligned}
\end{equation}
Since \(v\) is uniform on the unit sphere,
\begin{equation}
\mathbb E_v[vv^\top]=\frac{1}{D}I_D.
\end{equation}
Therefore, since \(P\) is a projection matrix satisfying \(P^\top=P\) and \(P^2=P\), we obtain
\begin{equation}
\begin{aligned}
\mathbb E_v\|PTv\|_2^2
&=
\operatorname{tr}
\!\left(
T^\top P T \,\mathbb E_v[vv^\top]
\right) \\
&=
\frac{1}{D}\operatorname{tr}(T^\top P T) \\
&=
\frac{1}{D}\|PT\|_F^2 \\
&=
\frac{1}{D}S(T).
\end{aligned}
\end{equation}
Thus,
\begin{equation}
\mathcal H(T)
=
\log K
-
\frac{1}{2KD}S(T)
+
\mathbb E_v[R(Tv)].
\end{equation}

It remains to bound the remainder uniformly in \(T\).
Let \(A\coloneq PT\). By Cauchy--Schwarz,
\begin{equation}
\mathbb E_v\|Av\|_2^3
\le
\sqrt{\mathbb E_v\|Av\|_2^2}
\sqrt{\mathbb E_v\|Av\|_2^4}.
\end{equation}
For \(v\sim \mathrm{Unif}(\mathbb S^{D-1})\), the standard fourth-moment identity gives
\begin{equation}
\begin{aligned}
\mathbb E_v\|Av\|_2^4
&=
\mathbb E_v\!\left[(v^\top A^\top A v)^2\right] \\
&\le
\frac{3\|A\|_F^4}{D(D+2)} \\
&\le
\frac{3\|A\|_F^4}{D^2},
\end{aligned}
\end{equation}
while
\begin{equation}
\mathbb E_v\|Av\|_2^2=\frac{\|A\|_F^2}{D}.
\end{equation}
Combining the two bounds yields
\begin{equation}
\begin{aligned}
\mathbb E_v\|PTv\|_2^3
&\le
\sqrt{3}\,
\frac{\|PT\|_F^3}{D^{3/2}} \\
&=
\sqrt{3}\,
\frac{S(T)^{3/2}}{D^{3/2}}.
\end{aligned}
\end{equation}
Moreover, since
\begin{equation}
S(T)=K-K\|\mu(T)\|_2^2,
\end{equation}
\(S(T)\) is bounded between \(0\) and \(K\). That is,
\begin{equation}
0\le S(T)\le K.
\end{equation}
Therefore,
\begin{equation}
\mathbb E_v\|PTv\|_2^3=O(D^{-3/2}),
\end{equation}
uniformly over all admissible text features \(T\), and hence
\begin{equation}
\mathcal H(T)
=
\log K
-
\frac{1}{2KD}S(T)
+
O(D^{-3/2}).
\end{equation}

Finally, we rewrite this expansion in terms of the corresponding regularizer.

For Eq.~\eqref{eq:ctpt_detail}, since
\(L_{\mathrm{disp}}(T)=-S(T)\),
we have
\begin{equation}
\mathcal H(T)
=
\frac{1}{2KD}L_{\mathrm{disp}}(T)
+
\log K
+
O(D^{-3/2}).
\end{equation}
Thus, in this case,
\begin{equation}
\alpha_{\mathrm{disp}}=\frac{1}{2KD},
\qquad
\beta_{\mathrm{disp}}=\log K.
\end{equation}

For Eq.~\eqref{eq:otpt_detail}, since
\[
L_{\mathrm{orth}}(T)
=
\frac{K}{2}\bigl(K-1-S(T)\bigr),
\]
we obtain
\begin{equation}
S(T)
=
K-1-\frac{2}{K}L_{\mathrm{orth}}(T).
\end{equation}
Substituting this into the entropy expansion gives
\begin{equation}
\begin{aligned}
\mathcal H(T)
&=
\frac{1}{K^2D}L_{\mathrm{orth}}(T) \\
&\quad+
\left(
\log K-\frac{K-1}{2KD}
\right)
+
O(D^{-3/2}).
\end{aligned}
\end{equation}
Thus, in this case,
\begin{equation}
\alpha_{\mathrm{orth}}=\frac{1}{K^2D},
\qquad
\beta_{\mathrm{orth}}
=
\log K-\frac{K-1}{2KD}.
\end{equation}

\begin{figure}[t!]
\vspace{-5pt}
\begin{center}
    \includegraphics[width=1.0\columnwidth]{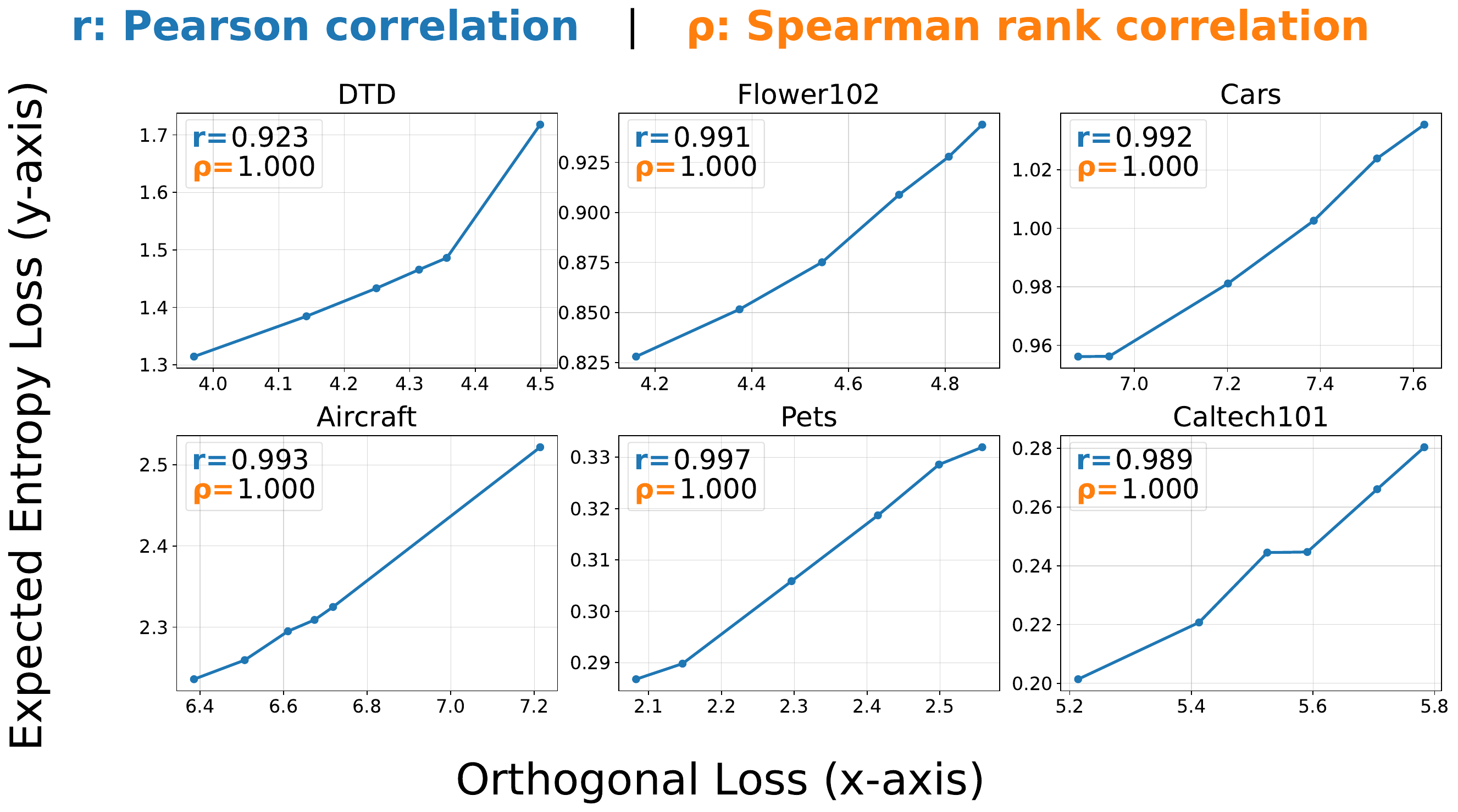}
    \vspace{-15pt}
    \caption{Correlation between expected EM loss and regularization loss.}
    \label{fig:correlation_em_reg}
\vspace{-18pt}
\end{center}
\end{figure}

Therefore, for either of the regularizers associated with Eq.~\eqref{eq:ctpt_detail} and Eq.~\eqref{eq:otpt_detail}, the expected EM loss satisfies
\begin{equation}
\mathcal H(T)
=
\alpha L_{\mathrm{reg}}^{\mathrm{th}}(T)
+
\beta
+
O(D^{-3/2}),
\qquad
\alpha>0.
\end{equation}
Hence, up to an \(O(D^{-3/2})\) remainder, the expected EM loss increases with the corresponding regularization loss.
This completes the proof. \(\square\)

Fig.~\ref{fig:correlation_em_reg} further supports this relationship across the settings used in our experiments.

\section{Connection Between the Proposed Flatness Loss and Sharpness of the Loss Landscape}
\label{sec:Proof_1}

Our goal is to clarify how the flatness loss \(\mathcal{L}_{\mathrm{flat}}\) in Eq.~\eqref{eq:flatness_loss} helps reduce the sharpness of other differentiable losses derived from the model output.
Specifically, we aim to show that constraining the deviation of the model output $f_{T}(C;\theta)$ under small perturbations effectively decreases the sharpness of any loss function defined in terms of the model output, $\ell(f_T(C,\theta))$, with respect to $(C,\theta)$.

To formalize this connection, we first analyze how small perturbations in $(C,\theta)$ affect the model output. In particular, we approximate $f_{T}(C+\varepsilon_{1};\,\theta+\varepsilon_{2})$ using a second-order Taylor expansion:
\begin{multline}
f_{T}(C+\varepsilon_{1};\,\theta+\varepsilon_{2}) \approx
f_{T}(C;\theta) + J_{C}\varepsilon_{1} + J_{\theta}\varepsilon_{2} \\
\shoveleft{+\tfrac{1}{2}\varepsilon_{1}^{\top}H_{CC}\varepsilon_{1}
+ \varepsilon_{1}^{\top}H_{C\theta}\varepsilon_{2}
+ \tfrac{1}{2}\varepsilon_{2}^{\top}H_{\theta\theta}\varepsilon_{2},}
\label{eq:taylor_expand}
\end{multline}
where \(J_{C}=\partial f_{T}/\partial C\) and \(J_{\theta}=\partial f_{T}/\partial\theta\) denote the Jacobians of the model output. \(H_{CC}\), \(H_{C\theta}\), and \(H_{\theta\theta}\) represent the corresponding Hessian blocks.

For a small deviation \(\Delta f_T = f_{T}(C+\varepsilon_{1},\,\theta+\varepsilon_{2}) - f_{T}(C,\theta)\), 
the flatness loss \(\mathcal{L}_{\mathrm{flat}}\) based on the cosine distance can be locally approximated as
\begin{equation}
\mathcal{L}_{\mathrm{flat}}
=\operatorname{dist}_{\text{cos}}(f_T+\Delta f_T,\,f_T)
\approx
\frac{1}{2\|f_T\|^{2}}\,\|\Delta f_T\|^{2},
\label{eq:cos_approx}
\end{equation}
where this approximation ignores higher-order terms and the dependence on the precise alignment between 
\(\Delta f_T\) and the normalized direction \(u = f_T/\|f_T\|\).
This local approximation shows that the flatness loss grows quadratically with the magnitude of the output deviation.

By substituting the Taylor expansion in Eq.~\eqref{eq:taylor_expand} into Eq.~\eqref{eq:cos_approx} and taking expectations over zero-mean isotropic Gaussian perturbations \(\varepsilon_{1},\varepsilon_{2}\), we obtain:
\begin{equation}
\begin{aligned}
\mathbb{E}_{\varepsilon_{1},\varepsilon_{2}}\!\big[\mathcal{L}_{\mathrm{flat}}\big]
\;\propto\;
&\underbrace{\|J_{C}\|^{2}_F}_{\text{input deviation}}
+\underbrace{\|J_{\theta}\|^{2}_F}_{\text{parameter deviation}}
+\underbrace{\|H_{C\theta}\|^{2}_F}_{\text{cross sharpness}} \\[2pt]
&+\underbrace{\|H_{CC}\|^{2}_F}_{\text{input sharpness}}
+\underbrace{\|H_{\theta\theta}\|^{2}_F}_{\text{parameter sharpness}}.
\end{aligned}
\label{eq:lflat_expectation}
\end{equation}
For simplicity, we ignore constant factors and higher-order terms and view Eq.~\eqref{eq:lflat_expectation} as capturing the dominant dependence of $\mathbb{E}[\mathcal{L}_{\mathrm{flat}}]$ on the derivatives of $f_T$.
This expression shows that the expectation of the flatness loss is proportional to the sum of the squared magnitudes of both first- and second-order derivatives of the model output. 
Consequently, $\mathcal{L}_{\mathrm{flat}}$ explicitly penalizes output-level sharpness by reducing excessive sensitivity with respect to both inputs $C$ and parameters $\theta$.

Let the scalar loss be defined as $\ell(f_T(C,\theta))=\phi(f_{T}(C;\theta),v)$, where $\phi:\mathbb{R}^{d}\!\to\!\mathbb{R}$ is a differentiable mapping and $v$ is arbitrary image features. Denote $g=\nabla_{f} \phi\!\big(f_{T}(C;\theta),v\big)$ as the gradient of $\ell$ with respect to the model output.
By applying the multivariate chain rule, the Hessian of $\ell$ with respect to $C$ and $\theta$ can be expressed as:
\begin{equation}
\begin{aligned}
\nabla^{2}_{CC}\ell \; &=\; J_{C}^{\!\top} H_{f} J_{C} \;+\; g^{\!\top}\! H_{CC},\\[2pt]
\nabla^{2}_{\theta\theta}\ell \; &=\; J_{\theta}^{\!\top} H_{f} J_{\theta} \;+\; g^{\!\top}\! H_{\theta\theta},
\end{aligned}
\end{equation}
where $H_{f}$ denotes the Hessian of the function $\phi$ with respect to the output $f_{T}$. 
The terms $J_{C}$, $J_{\theta}$, $H_{CC}$ and $H_{\theta\theta}$ correspond to the Jacobians and Hessians of $f_{T}(C;\theta)$ with respect to $C$ and $\theta$, as defined in Eq.~\eqref{eq:taylor_expand}.

Consequently, minimizing flatness loss $\mathcal{L}_{\mathrm{flat}}$---which reduces $\|J_{C}\|_{F}$, $\|J_{\theta}\|_{F}$, $\|H_{CC}\|_{F}$, and $\|H_{\theta\theta}\|_{F}$---tends to diminish the corresponding components of the Hessian of $\ell(f_T(C,\theta))$.
Because the Hessian reflects the sharpness of the loss surface, reducing it directly leads to a smoother landscape~\cite{Zhuang2022}.
Therefore, suppressing the sharpness of the model output inherently smooths the loss landscape, guiding optimization toward a flatter loss landscape.

\begin{figure}[t!]
\begin{center}
    \includegraphics[width=1.0\columnwidth]{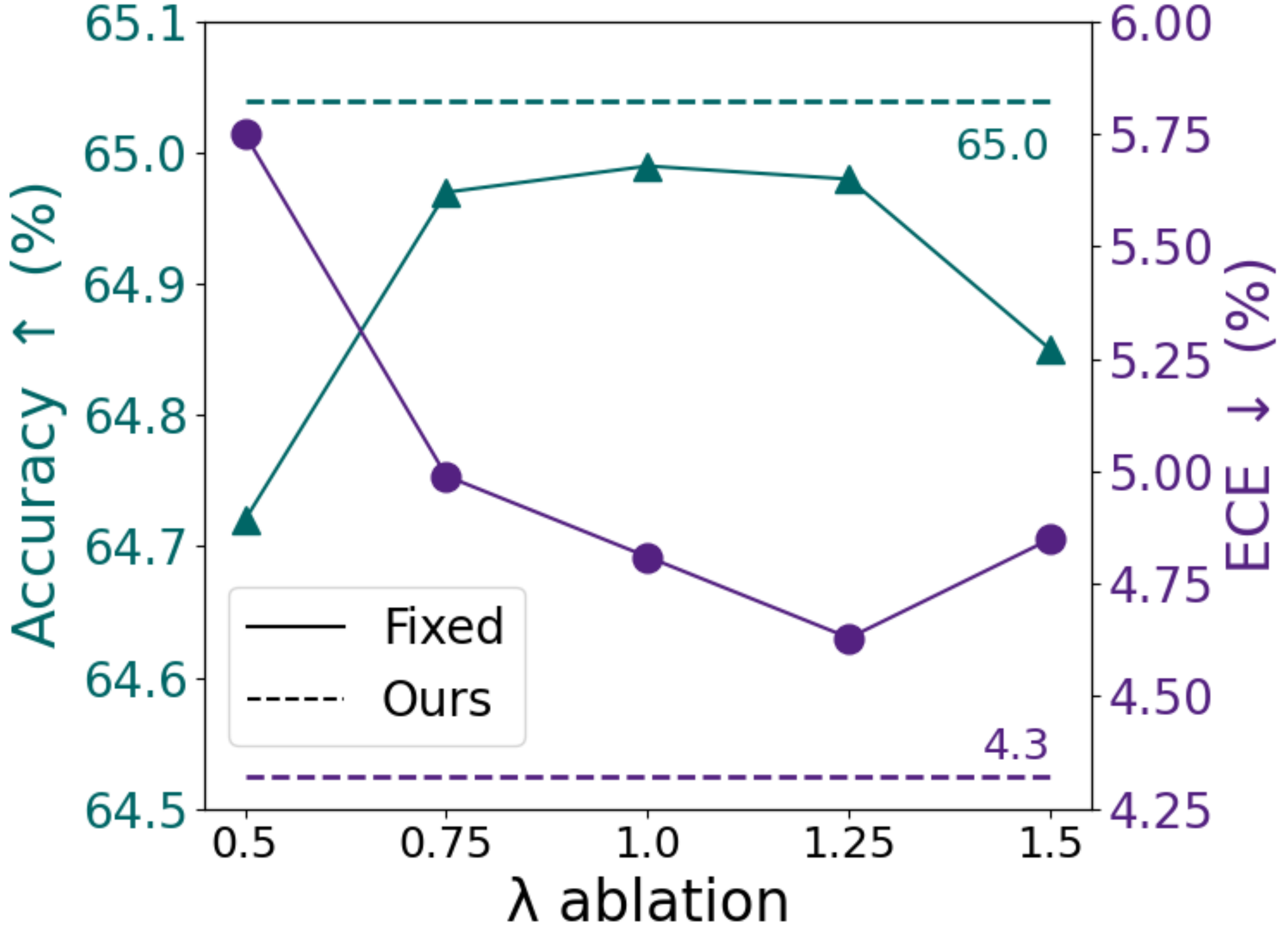}
    \vspace{-14pt}
    \caption{Comparison between fixed hyperparameter $\lambda$ and the proposed dynamic $\lambda$ strategy.}
    \label{fig:lambda_ablation}
\vspace{-20pt}
\end{center}
\end{figure}

\begin{figure*}[t!]
\begin{center}
\resizebox{0.9\linewidth}{!}{%
  \includegraphics[width=\textwidth]{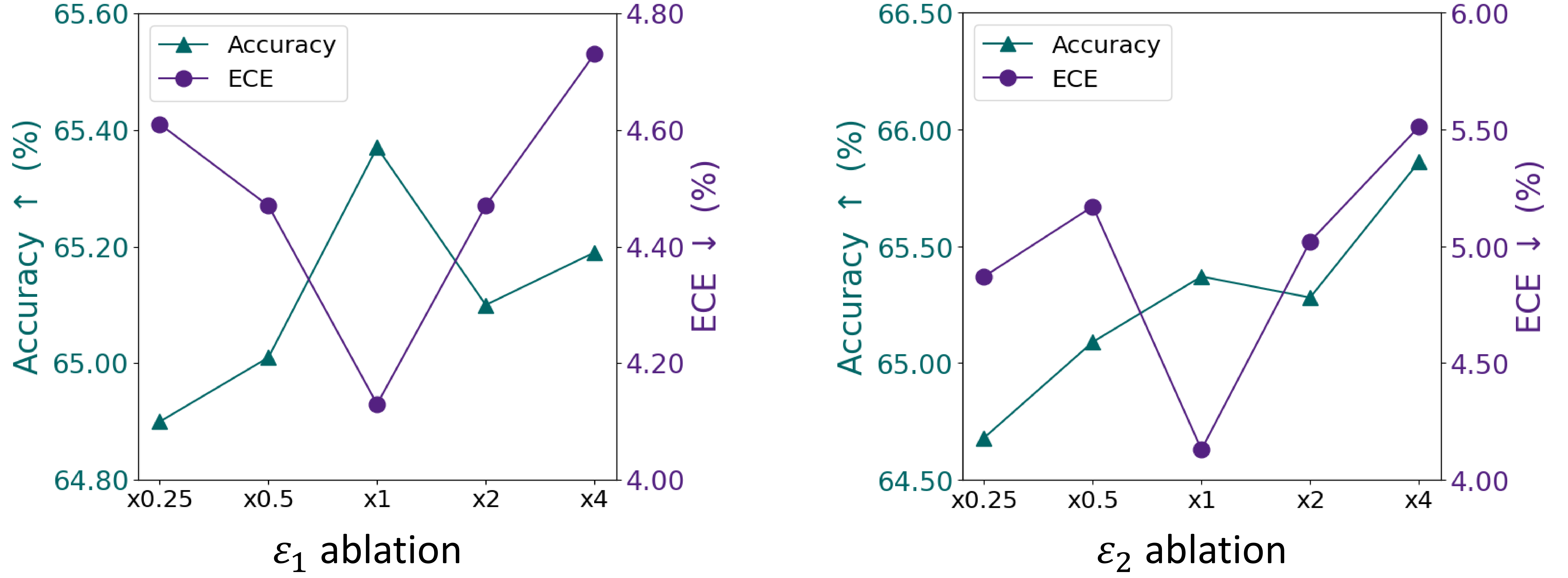}
}
\vspace{-8pt}
\caption{Ablation on perturbation magnitude. We scale the variances of the Gaussian distributions used to sample perturbations $\varepsilon_1$ and $\varepsilon_2$ and report the resulting performance. The results indicate that tuning the perturbation magnitude within an appropriate range is essential for achieving optimal performance.}
\label{fig:ablation_epsilon}
\vspace{-16pt}
\end{center}
\end{figure*}

\section{Hyperparameter Settings}
\label{sec:hyperparam}

\subsection{Analysis of Lambda}
In Eq.~\eqref{eq:ours_loss}, $\lambda$ scales the flatness loss $\mathcal{L}_{\mathrm{flat}}$ during prompt optimization.
Rather than using a fixed value of $\lambda$ across all datasets, we allow it to adapt automatically to dataset-specific characteristics.
Concretely, we set $\lambda = \gamma_1 + \frac{\gamma_2}{K}$ with $\gamma_1 = 1.0$ and $\gamma_2 = 0.15$, where $K$ denotes the number of classes.
This design reflects the intuition that preserving the original text features becomes more difficult on datasets with larger label spaces; accordingly, diminishing the contribution of the flatness loss helps maintain the underlying text semantics.
As shown in Fig.~\ref{fig:lambda_ablation}, this dataset-dependent scheduling of $\lambda$ consistently outperforms a fixed choice, yielding improvements in both accuracy and ECE across the fine-grained classification datasets.

\subsection{Analysis of the Magnitude of Perturbation}
In Fig.~\ref{fig:ablation_epsilon}, we conduct an ablation study on the magnitude of the two perturbations used in the flatness loss. 
Specifically, starting from our default setting where $\varepsilon_1$ and $\varepsilon_2$ are sampled from Gaussian distributions with variances $0.02$ and $0.005$, respectively, we scale these variances by constant factors and measure the resulting performance on the fine-grained classification datasets.
Here, a scaling factor of $\times 1$ corresponds to our default configuration.
Across both experiments, we observe a consistent trend: when the perturbation magnitudes are too small, the model exhibits lower accuracy and higher calibration error, while increasing $\epsilon_1$ and $\epsilon_2$ gradually improves both metrics.
However, once the perturbations exceed our default values, the calibration error starts to increase again and the overall performance degrades. 
This pattern indicates that choosing $\varepsilon_1$ and $\varepsilon_2$ within an appropriate range is crucial, and that our default setting yields a favorable trade-off between accuracy and calibration.

\begin{table}[t!]
  \centering
  \resizebox{1.0\columnwidth}{!}{%
    \begin{tabular}{%
      >{\centering\arraybackslash}p{0.28\linewidth}
      >{\centering\arraybackslash}p{0.28\linewidth}|
      >{\centering\arraybackslash}p{0.2\linewidth}
      >{\centering\arraybackslash}p{0.2\linewidth}}
      \toprule
      \textbf{Align Loss} &
      \textbf{Flat Loss} &
      \textbf{Acc.} &
      \textbf{ECE} \\
      \midrule
      Cosine & L2 & 64.70 & 4.87 \\
      Cosine & Cosine & 64.81 & 4.91 \\
      L2 & L2 & 64.56 & 4.80 \\
      \midrule
      \cellcolor{gray!20}L2 &
      \cellcolor{gray!20}Cosine &
      \cellcolor{gray!20}\textbf{65.37} &
      \cellcolor{gray!20}\textbf{4.13} \\
      \bottomrule
    \end{tabular}
  }
  \vspace{-8.0pt}
  \caption{Ablation study on distance functions used in the losses of the proposed pretraining stage. The method demonstrates robust performance, achieving stable accuracy and ECE regardless of the function selected.}
  \label{tab:ablation_dist}
  \vspace{-10.0pt}
\end{table}

\subsection{Analysis of Distance Function}
In our pretraining objective, we consider two choices of distance metrics for both the alignment loss in Eq.~\eqref{eq:align_loss} and the flatness loss in Eq.~\eqref{eq:flatness_loss}: L2 distance and cosine distance. Tab.~\ref{tab:ablation_dist} presents an ablation study in which we interchange these distance functions for each loss term while keeping all other components unchanged.
Across all configurations, our method consistently delivers higher accuracy than the baseline methods C-TPT and O-TPT, and also achieves lower ECE than C-TPT, demonstrating that the effectiveness of our approach is robust to the specific choice of distance metric.

\section{Datasets}
Following the baseline papers~\cite{Yoon2024, Sharifdeen2025}, we adopt the ten fine-grained classification datasets originally introduced in CoOp~\cite{Zhou2022b}.
These datasets span a broad set of visual domains, including plants and animals (Flower102, OxfordPets), textures (DTD), food imagery (Food101), and scenes (SUN397).
They also cover human action recognition (UCF101), satellite imagery analysis (EuroSAT), transportation categories such as cars and aircraft (StanfordCars, Aircraft), and the general-purpose dataset (Caltech101).

In addition, we evaluate on four ImageNet variants that are widely used to assess robustness under natural distribution shifts: ImageNet-V2, ImageNet-A, ImageNet-R, and ImageNet-Sketch. Compared to the standard ImageNet validation set, these datasets respectively contain re-collected samples, naturally adversarial images, artistic renditions (e.g., paintings and cartoons), and sketch drawings, thereby providing a diverse set of challenging test conditions for analyzing both accuracy and calibration.

\begin{table*}[t!]
\centering
\resizebox{1.0\textwidth}{!}{%
\begin{tabular}{l|c|cccccccccc|c}
\toprule
\textbf{Method} & \textbf{Metric} & \textbf{Air} & \textbf{Calt} & \textbf{Car} & \textbf{DTD} & \textbf{SAT} & \textbf{FLW} & \textbf{Food} & \textbf{Pets} & \textbf{SUN} & \textbf{UCF} & \textbf{Avg.} \\
\midrule
\multirow{3}{*}{O-TPT~\cite{Sharifdeen2025}} & AECE & 3.96 & 4.78 & 1.72 & 8.21 & 13.92 & 4.07 & 4.81 & 2.16 & 8.51 & 1.96 & 5.41 \\
                           & AURC & 0.57 & 0.01 & 0.14 & 0.31 & 0.40 & 0.10 & 0.05 & 0.02 & 0.18 & 0.14 & 0.19 \\
                           & MCE & 13.91 & 86.10 & 13.07 & 17.05 & 29.25 & 18.26 & 11.06 & 22.57 & 45.27 & 8.74 & 26.53 \\
\midrule
\cellcolor{gray!20} & \cellcolor{gray!20}AECE & \cellcolor{gray!20}7.52 & \cellcolor{gray!20}6.24 & \cellcolor{gray!20}2.16 & \cellcolor{gray!20}7.83 & \cellcolor{gray!20}5.36 & \cellcolor{gray!20}2.98 & \cellcolor{gray!20}1.98 & \cellcolor{gray!20}2.46 & \cellcolor{gray!20}3.20 & \cellcolor{gray!20}3.04 & \cellcolor{gray!20}\textbf{4.28} \\
\cellcolor{gray!20} & \cellcolor{gray!20}AURC & \cellcolor{gray!20}0.57 & \cellcolor{gray!20}0.01 & \cellcolor{gray!20}0.13 & \cellcolor{gray!20}0.30 & \cellcolor{gray!20}0.31 & \cellcolor{gray!20}0.10 & \cellcolor{gray!20}0.05 & \cellcolor{gray!20}0.02 & \cellcolor{gray!20}0.18 & \cellcolor{gray!20}0.12 & \cellcolor{gray!20}\textbf{0.18} \\
\multirow{-3}{*}{\cellcolor{gray!20}FPP (Ours)} & \cellcolor{gray!20}MCE & \cellcolor{gray!20}27.02 & \cellcolor{gray!20}62.33 & \cellcolor{gray!20}7.94 & \cellcolor{gray!20}20.03 & \cellcolor{gray!20}23.52 & \cellcolor{gray!20}11.80 & \cellcolor{gray!20}6.82 & \cellcolor{gray!20}24.78 & \cellcolor{gray!20}8.52 & \cellcolor{gray!20}10.80 & \cellcolor{gray!20}\textbf{20.36} \\
\bottomrule
\end{tabular}
} 
\vspace{-7.5pt}
\caption{Comparison of calibration metrics (AECE, AURC, MCE) between O-TPT and FPP using the CLIP-ViT/B16 backbone within the TPT framework under a predefined hard prompt (``a photo of a''). The best result for each metric is highlighted in \textbf{bold}.}
\label{tab:tpt_FG_additional}
\vspace{-6pt}
\end{table*}

\begin{table*}[t!]
\centering
\resizebox{1.0\textwidth}{!}{%
\begin{tabular}{l|c|cccccccccc|c}
\toprule
\textbf{Method} & \textbf{Metric} & \textbf{Air} & \textbf{Calt} & \textbf{Car} & \textbf{DTD} & \textbf{SAT} & \textbf{FLW} & \textbf{Food} & \textbf{Pets} & \textbf{SUN} & \textbf{UCF} & \textbf{Avg.} \\
\midrule
\multirow{2}{*}{O-TPT~\cite{Sharifdeen2025}} & SAM & 0.708 & 1.381 & 0.571 & 1.277 & 0.619 & 1.269 & 1.775 & 0.826 & 1.764 & 1.396 & 1.158 \\
 & Fisher-Rao & 0.871 & 0.452 & 1.939 & 0.920 & 0.255 & 0.279 & 1.163 & 0.035 & 0.862 & 0.213 & 0.699 \\
\midrule
\cellcolor{gray!20} & \cellcolor{gray!20}SAM & \cellcolor{gray!20}0.700 & \cellcolor{gray!20}0.659 & \cellcolor{gray!20}0.560 & \cellcolor{gray!20}0.956 & \cellcolor{gray!20}0.872 & \cellcolor{gray!20}0.748 & \cellcolor{gray!20}0.604 & \cellcolor{gray!20}0.487 & \cellcolor{gray!20}1.039 & \cellcolor{gray!20}0.796 & \cellcolor{gray!20}0.742 \\
\multirow{-2}{*}{\cellcolor{gray!20}FPP (Ours)} & \cellcolor{gray!20}Fisher-Rao & \cellcolor{gray!20}0.251 & \cellcolor{gray!20}0.048 & \cellcolor{gray!20}0.115 & \cellcolor{gray!20}0.228 & \cellcolor{gray!20}0.244 & \cellcolor{gray!20}0.123 & \cellcolor{gray!20}0.049 & \cellcolor{gray!20}0.057 & \cellcolor{gray!20}0.130 & \cellcolor{gray!20}0.122 & \cellcolor{gray!20}\textbf{0.137} \\
\bottomrule
\end{tabular}
}
\vspace{-7.5pt}
\caption{Comparison of flatness measures between O-TPT and FPP using the CLIP-ViT/B16 backbone within the TPT framework under a predefined hard prompt (``a photo of a''). Lower values indicate a flatter loss landscape.}
\label{tab:fisher_rao_flatness}
\vspace{-8pt}
\end{table*}

\section{Calibration Metrics}
\label{sec:metric}
Following previous works~\cite{Yoon2024, Sharifdeen2025}, we evaluate calibration using two metrics: (1) Expected Calibration Error (ECE)~\cite{Naeini2015} and Static Calibration Error (SCE)~\cite{Jeremy2019}.
Specifically, ECE measures the discrepancy between predicted confidence and accuracy by partitioning predictions into $M$ bins:
\begin{equation}
\text{ECE} = \sum_{b=1}^{B} \frac{|A_b|}{M} \Big| \text{acc}(A_b) - \text{conf}(A_b) \Big|,
\end{equation}
where $B$ is the number of bins used to divide the prediction confidences, $A_b$ denotes the set of samples whose confidence scores fall into the $b$-th bin, and $|A_b|$ indicates the number of samples in $A_b$.
$M$ is the total number of predictions, $\text{acc}(\cdot)$ represents the prediction accuracy, and $\text{conf}(\cdot)$ is the average confidence of the samples.
In addition, the SCE extends this idea to multi-class settings by computing calibration error across both confidence bins and classes:
\begin{equation}
\text{SCE} = \frac{1}{K} \sum_{k=1}^{K} \sum_{b=1}^{B} \frac{|A_{k,b}|}{M}\Big| \text{acc}(A_{k,b}) - \text{conf}(A_{k,b}) \Big|.
\end{equation}
Here, $A_{k,b}$ denotes the set of samples belonging to class $k$ whose prediction confidence falls into the $b$-th bin.

\section{Results on Additional Calibration Metrics}
We further evaluated our method on a broader set of calibration metrics beyond ECE and SCE, as shown in Tab~\ref{tab:tpt_FG_additional}. 
Specifically, we compared our method against O-TPT on AECE, AURC, and MCE. 
AECE~\cite{Mukhoti2020} provides a more robust estimate of calibration by reducing dependence on a fixed binning scheme. 
AURC~\cite{Geifman2019} captures the overall risk-coverage trade-off when predictions are ranked by confidence. 
MCE~\cite{Naeini2015} measures the largest calibration error across bins. 
Our method consistently achieved better results than O-TPT across all three metrics.

\begin{table*}[t!]
\centering
\resizebox{0.7\textwidth}{!}{%
\begin{tabular}{l|c|ccccc|c}
\toprule
\textbf{Method} & \textbf{Metric} & \textbf{Calt} & \textbf{Car} & \textbf{DTD} & \textbf{FLW} & \textbf{Food} & \textbf{Avg.} \\
\midrule
\multirow{2}{*}{C-TPT}          & Std. Acc. & 0.12 & 0.16 & 0.16 & 0.22 & 0.20 & 0.17 \\
                                & Std. ECE  & 0.24 & 0.18 & 0.24 & 0.12 & 0.19 & 0.194 \\
\midrule
\multirow{2}{*}{O-TPT}          & Std. Acc. & 0.14 & 0.11 & 0.03 & 0.10 & 0.19 & \textbf{0.11} \\
                                & Std. ECE  & 0.17 & 0.25 & 0.14 & 0.20 & 0.10 & 0.177 \\
\midrule
\cellcolor{gray!20}      
                                & \cellcolor{gray!20}Std. Acc. & \cellcolor{gray!20}0.05 & \cellcolor{gray!20}0.05 & \cellcolor{gray!20}0.27 & \cellcolor{gray!20}0.10 & \cellcolor{gray!20}0.18 & \cellcolor{gray!20}0.13 \\
\cellcolor{gray!20}                                
\multirow{-2}{*}{\cellcolor{gray!20}FPP (Ours)}
                                & \cellcolor{gray!20}Std. ECE  & \cellcolor{gray!20}0.12 & \cellcolor{gray!20}0.27 & \cellcolor{gray!20}0.15 & \cellcolor{gray!20}0.21 & \cellcolor{gray!20}0.04 & \cellcolor{gray!20}\textbf{0.158} \\
\bottomrule
\end{tabular}
} 
\vspace{-7.5pt}
\caption{Standard deviation across three different seed runs.}
\label{tab:std_FG}
\vspace{-5.0pt}
\end{table*}

\section{Additional Sharpness Analysis}
To further support our claim that the proposed method converges to flatter minima, we provide an additional sharpness analysis in Tab.~\ref{tab:fisher_rao_flatness}. 
Specifically, under the TPT framework, we report the Fisher-Rao norm~\cite{Liang2019} for both O-TPT and FPP, along with SAM-based flatness.
The results show that FPP consistently achieves smaller values than O-TPT across all datasets.

\section{Standard Deviation Across Different Seeds}
In Tab.~\ref{tab:std_FG}, following the evaluation protocol of O-TPT, we report the standard deviation of accuracy and ECE across five fine-grained datasets.
For three different random seeds, we independently perform both the pretraining and test-time adaptation (TTA) stages, compute the standard deviation for each dataset, and then report the mean standard deviation across datasets.
Specifically, we compute the standard deviation over three different random seeds for both the pretraining and TTA stages, and then report the mean standard deviation across datasets.
Our method exhibits an accuracy standard deviation comparable to O-TPT, while attaining the lowest ECE deviation among all methods.
These results indicate that our approach not only achieves strong average performance but also maintains consistently stable behavior across multiple runs with different random initializations.